\documentclass{article}

\usepackage{microtype}
\usepackage{graphicx}
\usepackage{subfigure}
\usepackage{booktabs} 

\usepackage{hyperref}

\usepackage[accepted]{icml2023}

\usepackage{amsmath}
\usepackage{amssymb}
\usepackage{mathtools}
\usepackage{amsthm}
\usepackage{colortbl, xcolor}
\usepackage[capitalize,noabbrev]{cleveref}

\usepackage{multirow}

\theoremstyle{plain}
\newtheorem{theorem}{Theorem}[section]
\newtheorem{proposition}[theorem]{Proposition}

\theoremstyle{definition}

\theoremstyle{remark}

\usepackage[textsize=tiny]{todonotes}

\usepackage{caption}
\icmltitlerunning{CoDi: Co-evolving Contrastive Diffusion Models for Mixed-type Tabular Synthesis}

\begin{document}

\twocolumn[
\icmltitle{CoDi: Co-evolving Contrastive Diffusion Models\\ for Mixed-type Tabular Synthesis}



\icmlsetsymbol{equal}{*}

\begin{icmlauthorlist}
\icmlauthor{Chaejeong Lee}{equal,yyy}
\icmlauthor{Jayoung Kim}{equal,yyy}
\icmlauthor{Noseong Park}{yyy}
\end{icmlauthorlist}

\icmlaffiliation{yyy}{Department of Artificial Intelligence, Yonsei University, Seoul, South Korea}

\icmlcorrespondingauthor{Noseong Park}{noseong@yonsei.ac.kr}

\icmlkeywords{Machine Learning, ICML}

\vskip 0.3in
]



\printAffiliationsAndNotice{\icmlEqualContribution} 
\begin{abstract}
With growing attention to tabular data these days, the attempt to apply a synthetic table to various tasks has been expanded toward various scenarios. Owing to the recent advances in generative modeling, fake data generated by tabular data synthesis models become sophisticated and realistic. However, there still exists a difficulty in modeling discrete variables (columns) of tabular data. In this work, we propose to process continuous and discrete variables separately (but being conditioned on each other) by two diffusion models. The two diffusion models are co-evolved during training by reading conditions from each other. In order to further bind the diffusion models, moreover, we introduce a contrastive learning method with a negative sampling method. In our experiments with 11 real-world tabular datasets and 8 baseline methods, we prove the efficacy of the proposed method, called \texttt{CoDi}. Our code is available at \url{https://github.com/ChaejeongLee/CoDi}.

\end{abstract}

\section{Introduction}\label{intro}
Tabular data is one of the most frequently occurring data types in real-world applications. Considering that tabular data attracts much attention these days \cite{borisov2021deep, shwartz2022tabular, yin2020tabert, gorishniy2021revisiting}, synthesizing tabular data is a timely research issue. The quality of generated tabular data has significantly improved owing to recent advancements in the generative model paradigm~\cite{tablegan,ctgan,octgan,itgan,stasy}.

In spite of the advancement, there still exists a challenging issue in the current state-of-the-art tabular data synthesis methods. To be specific, tabular data usually consists of mixed data types, i.e., continuous and discrete variables. Due to the complicated nature of tabular data, pre/post-processing of the tabular data is inevitable, and the performance of tabular data synthesis methods is highly dependent on the pre/post-processing method. While various processing methods for continuous variables have been proposed, e.g., standardizing continuous values using variational Gaussian mixture models~\cite{ctgan} and applying the logarithm transformation to treat long-tailed distributions~\cite{ctabgan}, handling discrete variables remains challenging.
The most common way to treat discrete variables is to sample in continuous spaces after their one-hot encoding (and sometimes via Gumbel-softmax)~\cite{ctgan, octgan, sos, stasy}, which may lead to sub-optimal results due to sampling mistakes.


Moreover, treating discrete variables in continuous spaces is also problematic in terms of the entire learned data distribution. When continuous and discrete variables are processed in a same manner, it is likely that inter-column correlations --- in particular, the correlation between continuous and discrete variables --- are compromised in the learned distribution. Therefore, we are interested in processing continuous and discrete variables in more robust ways.

\begin{figure}[t]
        \centering
        \includegraphics[width=0.7\columnwidth]{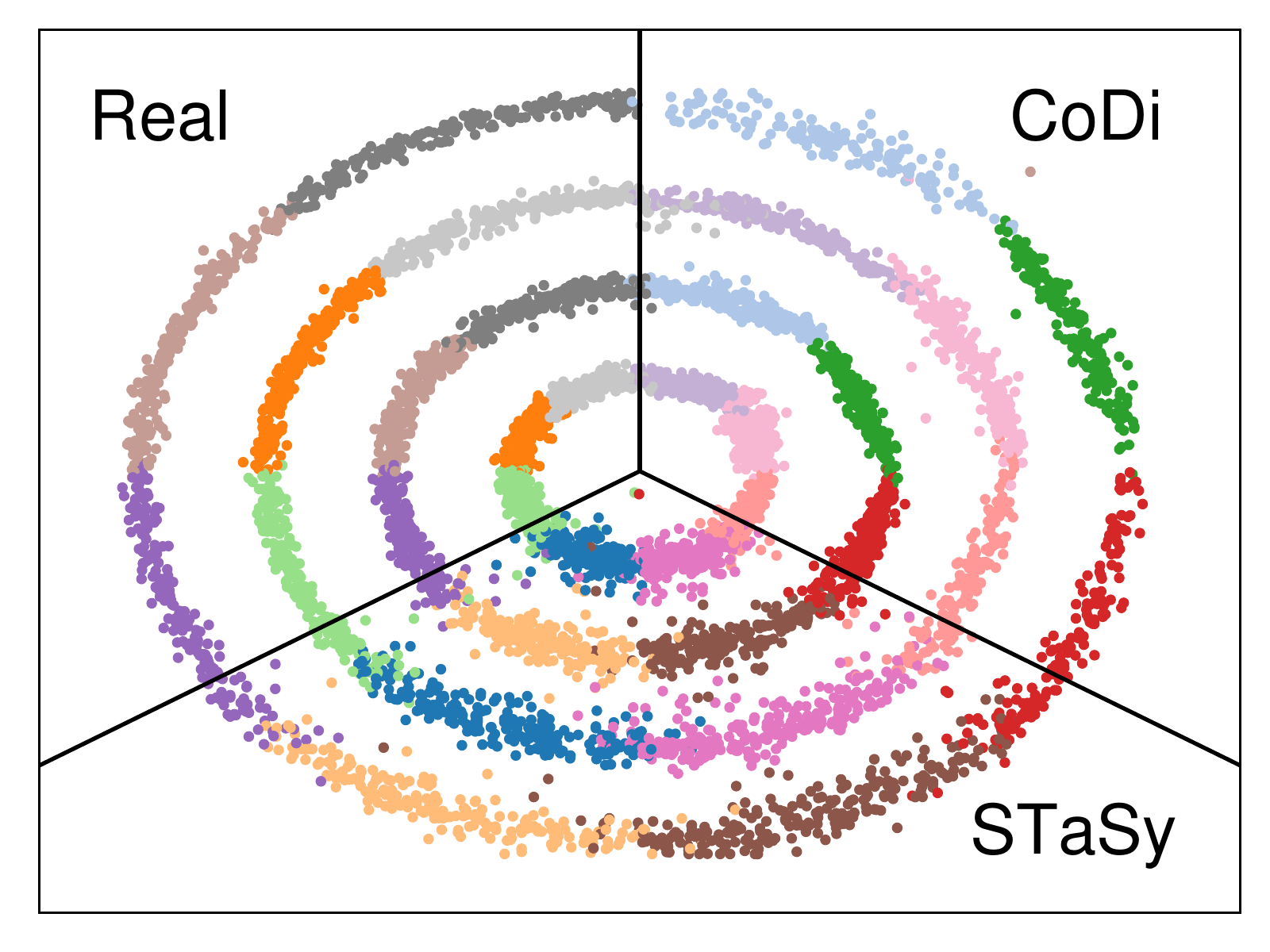}
        \caption{Preliminary experiment on a toy dataset. The dataset contains 4 columns, which are two continuous (the x-axis and y-axis) and two discrete (16 colors and 4 circles) columns. (Left) is a scatter plot of real data, (Bottom) is synthesized data by \texttt{STaSy}~\cite{stasy}, and (Right) is synthesized data by our proposed method. Detailed information and visualizations are in Appendix~\ref{sec:apd_toy}.}
        \label{fig:toydata}
\vspace{-0.5em}
\end{figure}

In this paper, we propose a technique for tabular data synthesis that incorporates two diffusion models to handle continuous and discrete variables. To be specific, one diffusion model for continuous variables works in a continuous space, and the other works in a discrete space.
 For better learning the mixed data distribution, our proposed method contains two design points: i) co-evolving conditional diffusion models, and ii) contrastive training for better connecting them.
 In our description below, let $\mathbf{x}_0 = (\mathbf{x}_0^C, \mathbf{x}_0^D)$, which consists of continuous and discrete values, be a sample (or record) of tabular data. $\mathbf{x}_t = (\mathbf{x}_t^C, \mathbf{x}_t^D)$ means its diffused sample at time (or step) $t$.

\noindent \textbf{Co-evolving conditional diffusion models} To make the two diffusion models able to synthesize one tabular data, we make them read conditions from each other as in Fig.~\ref{fig:method1}. The two models simultaneously perturb continuous and discrete variables at each forward step. In detail, the continuous (resp. discrete) model reads the perturbed discrete (resp. continuous) sample as a condition at the same time step. 
For the reverse process of the continuous (resp. discrete) diffusion model, the model denoises the sample $\mathbf{x}_t^C$ (resp. $\mathbf{x}_t^D$) conditioned both on the continuous sample $\mathbf{x}_{t+1}^C$ and discrete sample $\mathbf{x}_{t+1}^D$ from its previous step.



\noindent \textbf{Contrastive learning for tabular data}
To bind the two conditional diffusion models further, we design a contrastive learning method. Our contrastive learning process is applied to the continuous and discrete diffusion models separately. We also design a negative sampling method for tabular data, which focuses on defining a negative condition that permutes the pair of continuous and discrete variable sets. For simplicity but without loss of generality, we describe a process only for the continuous diffusion model (from which the contrastive learning for the discrete diffusion can be easily deduced). Given an anchor sample $\mathbf{x}_0^C$, we generate a continuous positive sample $\hat{\mathbf{x}}_0^{C+}$ from a continuous diffusion model conditioned on $\mathbf{x}_0^D$. For a negative sample $\hat{\mathbf{x}}_0^{C-}$, we randomly permute the condition parts, and therefore, negative condition $\mathbf{x}_0^{D-}$ is an inappropriate counterpart for $\mathbf{x}_0^C$. As a result, we generate $\hat{\mathbf{x}}_0^{C-}$ conditioned on $\mathbf{x}_0^{D-}$.


The two proposed design points for tabular data synthesis considerably improve the sampling quality over state-of-the-art methods. 
As shown in Fig.~\ref{fig:toydata}, our method demonstrates its overall efficacy, where Fig.~\ref{fig:toydata} visualizes sampling outcomes with synthetic toy tabular data. In Sec.~\ref{sec:experiments}, we provide more details with our experiments on 11 datasets and 8 baselines showing that our method effectively models real-world tabular data distribution by processing mixed-type tabular data in the proper spaces. 

We summarize the contributions of this paper as follows:
\begin{enumerate}
    \item We propose to separately train two diffusion models for continuous and discrete variables which consist of tabular data, to better learn their own distributions. To our knowledge, we are the first proposing separately processing them with two physically separated models.
    \item To bridge the two diffusion models, however, we propose to design them using co-evolving conditional continuous and discrete diffusion models, which are conditioned on each other. Moreover, we also design a contrastive learning method to reinforce the binding between the models further.
    \item 
    Consequently, \textit{the generative learning trilemma}~\cite{tackling}, which is the 3 key requirements for generative models including the sampling quality, diversity, and sampling time, has been improved over state-of-the-art methods, as shown in Sec.~\ref{sec:experiments}. 
\end{enumerate}

\section{Background}
\subsection{Diffusion Probabilistic Models}
Diffusion probabilistic models~\cite{sohl2015deep} are deep generative models defined from a forward and reverse Markov process. The forward process is to gradually corrupt a sample $\mathbf{x}_0$ into a noisy vector $\mathbf{x}_T$, as follows: \begin{equation}\label{eq:forward}
q(\mathbf{x}_{1:T}|\mathbf{x}_0) \coloneqq \prod_{t=1}^T{q(\mathbf{x}_{t}|\mathbf{x}_{t-1})}.
\end{equation} 
The reverse process is to remove noises and generate a fake sample $\hat{\mathbf{x}}_0$  from $\mathbf{x}_T$, as follows: \begin{equation}
p_{\theta}(\mathbf{x}_{0:T}) \coloneqq p(\mathbf{x}_{T})\prod_{t=1}^T{p_\theta(\mathbf{x}_{t-1}|\mathbf{x}_{t})},
\end{equation} where $p_\theta(\mathbf{x}_{t-1}|\mathbf{x}_{t})$ represents the reverse of the forward transition probability, approximated by a neural network. The parameter $\theta$ optimizes a variational upper bound on the negative log-likelihood: \begin{equation}\label{eq:loss_dis}
\begin{aligned}
L_{\mathrm{vb}} &= \mathbb{E}_q\bigg[\underbrace{D_{\mathrm{KL}}[q(\mathbf{x}_T|\mathbf{x}_0) || p(\mathbf{x}_T)]}_{L_T} \underbrace{-\log{p_\theta(\mathbf{x}_0|\mathbf{x}_1)}}_{L_0} \\
&+ \sum_{t=2}^T\underbrace{{D_{\mathrm{KL}}(q(\mathbf{x}_{t-1}|\mathbf{x}_t, \mathbf{x}_0)||p_\theta(\mathbf{x}_{t-1}|\mathbf{x}_t))}}_{L_{t-1}}\bigg].
\end{aligned}
\end{equation}

\subsubsection{Continuous Spaces}\label{sec:continuous}
The diffusion process in continuous spaces~\cite{ho2020denoising} defines the forward and reverse transition probabilities with a prior distribution $p(\mathbf{x}_T)=\mathcal{N}(\mathbf{x}_T;\mathbf{0},\mathbf{I})$, as follows: \begin{equation}\label{eq:forward_con}
q(\mathbf{x}_{t}|\mathbf{x}_{t-1})= \mathcal{N}(\mathbf{x}_t;\sqrt{1-\beta_t}\mathbf{x}_{t-1}, \beta_t\mathbf{I}),
\end{equation}
\begin{equation}\label{eq:reverse_con}
p_\theta(\mathbf{x}_{t-1}|\mathbf{x}_{t})=\mathcal{N}(\mathbf{x}_{t-1};\boldsymbol {\mu}_\theta(\mathbf{x}_t, t), \boldsymbol{\Sigma}_\theta(\mathbf{x}_t, t)), 
\end{equation} respectively, where gaussian noise is added to the sample according to a variance schedule $\beta_t\in(0,1)$.

The simplified objective function for approximating $p_\theta(\mathbf{x}_{t-1}|\mathbf{x}_{t})$ is defined as: \begin{equation}\label{eq:loss_con}
L_{\mathrm{simple}}(\theta)\coloneqq \mathbb{E}_{t,\mathbf{x}_0,\boldsymbol{\epsilon}}\Big[\big\Vert\boldsymbol{\epsilon}-\boldsymbol{\epsilon}_{\theta}(\mathbf{x}_t, t)\big\Vert^2\Big],
\end{equation}
where $\boldsymbol{\epsilon}_\theta$ is a neural network, which predicts noise $\boldsymbol{\epsilon}\sim\mathcal{N}(\mathbf{0},\mathbf{I})$. Using the predicted noise, one can generate a fake sample $\hat{\mathbf{x}}_0$.
\subsubsection{Discrete Spaces}\label{sec:discrete}

To handle discrete data, e.g., text and images~\cite{hoogeboom2021argmax,austin2021structured}, the diffusion process can be defined in discrete spaces using categorical distributions.

The forward and reverse transition probabilities in discrete spaces are as follows: \begin{equation}\label{eq:forward_dis}
q(\mathbf{x}_{t}|\mathbf{x}_{t-1})= \mathcal{C}(\mathbf{x}_t;(1-\beta_t)\mathbf{x}_{t-1} + \beta_t/K),
\end{equation}
\begin{equation}\label{eq:reverse_dis}
p_\theta(\mathbf{x}_{t-1}|\mathbf{x}_{t})=\sum_{\hat{\mathbf{x}}_0=1}^Kq(\mathbf{x}_{t-1}|\mathbf{x}_t,\hat{\mathbf{x}}_0)p_\theta(\hat{\mathbf{x}}_0|\mathbf{x}_t),
\end{equation}
where $\mathcal{C}$ indicates a categorical distribution, $K$ is the number of categories, and uniform noise is added to the sample according to $\beta_t$. The forward process posterior $q(\mathbf{x}_{t-1}|\mathbf{x}_t,\mathbf{x}_0)$ can be expressed as follows: \begin{equation}
    q(\mathbf{x}_{t-1}|\mathbf{x}_t,\mathbf{x}_0)=\frac{q(\mathbf{x}_t|\mathbf{x}_{t-1},\mathbf{x}_0)q(\mathbf{x}_{t-1}|\mathbf{x}_0)}{q(\mathbf{x}_t|\mathbf{x}_0)},
\end{equation}
which allows the closed-form computation of the KL divergence in $L_{t-1}$ of Eq.~\eqref{eq:loss_dis}.

\subsection{Tabular Data Synthesis}

There are many tabular data synthesis methods to create realistic fake tables for various purposes. For instance, \citet{7796926} utilizes a recursive table modeling with a Gaussian copula for synthesizing continuous variables. On the other hand, Bayesian networks~\cite{10.1145/3134428, avino2018generating} and decision trees~\cite{article} can be used for discrete variables. 
With great advancement in generative modeling, there exists an attempt to synthesize tabular data using GANs. 
 \texttt{TableGAN}~\cite{tablegan} utilizes convolutional neural networks to improve the quality of synthesized tabular data and prediction on label column accuracy. \texttt{CTGAN} and \texttt{TVAE}~\cite{ctgan} propose a column-type-specific pre-processing method to deal with the challenges in tabular data, for which tabular data usually consists of mixed-type variables and the variables follow multi-modal distributions.  In specific, they approximate the discrete variables to the continuous spaces by using Gumbel-Softmax. \texttt{OCT-GAN}~\cite{octgan} is a generative model based on neural ODEs. SOS~\cite{sos} and \texttt{STaSy}~\cite{stasy} are state-of-the-art tabular data synthesis methods, which are based on the score-based generative regime. The former focuses on synthesizing minority class(es) in classification data, while the latter generates the entire data.

\section{Proposed Method}
In Sec.~\ref{sec:co-evolving}, we introduce two diffusion models, one for continuous, and the other for discrete variables, and combine them to design co-evolving conditional diffusion models. Then, we present a contrastive learning method to improve the connection between the models for the two types of variables as in Sec.~\ref{sec:contrastive}. Our network architecture modified from U-Net~\cite{ronneberger2015u} is in Appendix~\ref{sec:net_archi}.
\begin{figure}[t]
        \centering
        \includegraphics[width=0.98\linewidth]{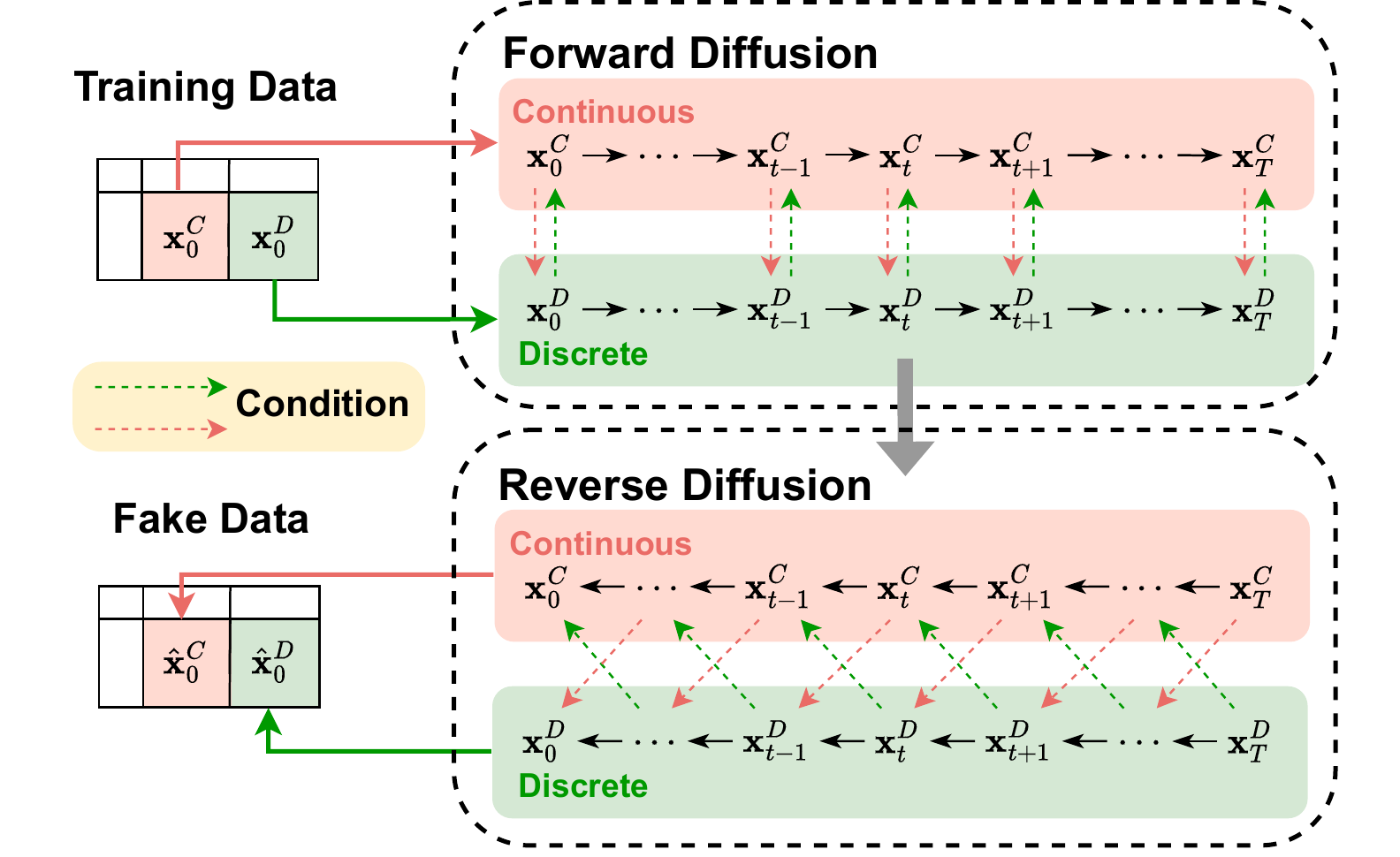}
        \caption{The overall workflow of co-evolving conditional diffusion models. Two diffusion models are connected by reading conditions from each other.}
        \label{fig:method1}
\vspace{-1em}
\end{figure}

\subsection{Co-evolving Conditional Diffusion Models}\label{sec:co-evolving}

Fig.~\ref{fig:method1} shows an overall workflow of the co-evolving conditional diffusion models. Given a sample $\mathbf{x}_0$ which consists of mixed types of variables, we assume without loss of generality that $\mathbf{x}_0$ contains $N_C$ continuous columns $C=\{C_1, C_2, \dots, C_{N_C}\}$ and $N_D$ discrete columns $D=\{D_1, D_2, \dots, D_{N_D}\}$, where $\mathbf{x}_0 = (\mathbf{x}_0^C, \mathbf{x}_0^D)$.

To synthesize samples from the space to which each type belongs, we train two diffusion models for the two variable types separately --- however, the two diffusion models read conditions from each other since their diffusion/denoising processes are intercorrelated. We call them as \emph{continuous} and \emph{discrete} diffusion models, respectively. When training for continuous variables $\mathbf{x}_0^C$, we use the method described in Sec.~\ref{sec:continuous}. For discrete variables $\mathbf{x}_0^D$, the method in Sec.~\ref{sec:discrete} is used.

To generate one related data pair with two models, we input each other's output as a condition. The pair $(\mathbf{x}_0^C, \mathbf{x}_0^D)$ are then simultaneously perturbed at each forward time step $t$ in each space. 
To be specific, $\mathbf{x}_t^C$, which is the perturbed data in the continuous diffusion model, will be the condition of $\mathbf{x}_t^D$ in the discrete diffusion model, and vice versa.
The model parameter $\theta_{C}$ (resp. $\theta_{D}$) is updated based on the following equations: \begin{equation}\label{eq:diff_con}
L_{\mathrm{Diff_C}}(\theta_C)\coloneqq \mathbb{E}_{t,\mathbf{x}_0^C,\boldsymbol{\epsilon}}\Big[\big\Vert\boldsymbol{\epsilon}-\boldsymbol{\epsilon}_{\theta_C}(\mathbf{x}_t^C, t \mid \mathbf{x}_t^D)\big\Vert^2\Big],
\end{equation}
\begin{equation}\label{eq:diff_dis}
\resizebox{0.89\hsize}{!}{%
$\begin{aligned}
L_{\mathrm{Diff_D}}(\theta_D) &= \mathbb{E}_q\bigg[\underbrace{D_{\mathrm{KL}}[q(\mathbf{x}_T^D|\mathbf{x}_0^D) || p(\mathbf{x}_T^D)]}_{L_T} \underbrace{-\log{p_{\theta_D}(\mathbf{x}^D_0|\mathbf{x}^D_1, \mathbf{x}_1^C})}_{L_0} \\
&+ \sum_{t=2}^T\underbrace{{D_{\mathrm{KL}}(q(\mathbf{x}_{t-1}^D|\mathbf{x}_t^D, \mathbf{x}_0^D)||p_{\theta_D}(\mathbf{x}_{t-1}^D|\mathbf{x}_t^D, \mathbf{x}_t^C))}}_{L_{t-1}}\bigg], 
\end{aligned}$
}\end{equation}
where $L_{\mathrm{Diff_C}}(\theta_C)$ and $L_{\mathrm{Diff_D}}(\theta_D)$ are the loss functions for the continuous and discrete diffusion models, respectively.

For the reverse process, generated samples, $\hat{\mathbf{x}}_0^C$ and $\hat{\mathbf{x}}_0^D$, are progressively synthesized from each noise space. The prior distributions of the two models are $p(\mathbf{x}_T^C)=\mathcal{N}(\mathbf{x}_T^C;\mathbf{0},\mathbf{I})$ and $p(\mathbf{x}_T^{D_i})=\mathcal{C}(\mathbf{x}_T^{D_i};{1/{K_i}})$, where $\{K_i\}_{i=1}^{N_D}$ is the number of categories of the discrete column $\{D_i\}_{i=1}^{N_D}$. 
After sampling noisy vectors, the two diffusion models convert the noises into fake samples while being conditioned on the denoised samples at the previous time step.
In detail, to denoise one step from $\mathbf{x}_{t+1}^C$ (resp. $\mathbf{x}_{t+1}^D$) to $\mathbf{x}_{t}^C$ (resp. $\mathbf{x}_{t}^D$), 
the continuous diffusion model (resp. the discrete diffusion model)
is conditioned both on the continuous sample $\mathbf{x}_{t+1}^C$ and discrete sample $\mathbf{x}_{t+1}^D$, which allows the collaboration of the two models to generate a related data record $\hat{\mathbf{x}}_0 = (\hat{\mathbf{x}}_0^C, \hat{\mathbf{x}}_0^D$).

\begin{proposition}\label{prop:cond}
The two forward processes of the continuous and discrete diffusion models are defined as follows:
\begin{equation}
q(\mathbf{x}_{t}^C|\mathbf{x}_0^C)= \mathcal{N}(\mathbf{x}_t^C;\sqrt{\bar{\alpha}_t}\mathbf{x}_{0}^C, (1-\bar{\alpha}_t)\mathbf{I}),
\end{equation}
\begin{equation}
q(\mathbf{x}_{t}^{D_i}|\mathbf{x}_0^{D_i})= \mathcal{C}(\mathbf{x}_t^{D_i};\bar{\alpha}_t\mathbf{x}_{0}^{D_i}+(1-\bar{\alpha}_t)/K_i),
\end{equation}
where $ 1\leq i \leq N_D$, $\alpha_t\coloneqq1-\beta_t$ and $\bar{\alpha}_t\coloneqq\prod_{i=1}^t{\alpha_i}$.

In addition, the reverse processes of the co-evolving conditional diffusion models are defined as follows:
\begin{equation}
p_{\theta_C}(\mathbf{x}_{0:T}^C) \coloneqq p(\mathbf{x}_{T}^C)\prod_{t=1}^T{p_{\theta_C}(\mathbf{x}_{t-1}^C|\mathbf{x}_{t}^C, \mathbf{x}_{t}^D)},
\end{equation}
\begin{equation}
p_{\theta_D}(\mathbf{x}_{0:T}^{D_i}) \coloneqq p(\mathbf{x}_{T}^{D_i})\prod_{t=1}^T{p_{\theta_D}(\mathbf{x}_{t-1}^{D_i}|\mathbf{x}_{t}^{D_i}, \mathbf{x}_{t}^C)}, 
\end{equation}
where $1\leq i \leq N_D$ and the reverse transition probabilities are defined in Appendix~\ref{sec:apd_proposition}.
\end{proposition}

\begin{figure}[t]
        \centering
        \includegraphics[trim={0cm 0cm 1cm 0cm},clip, width=1\columnwidth]{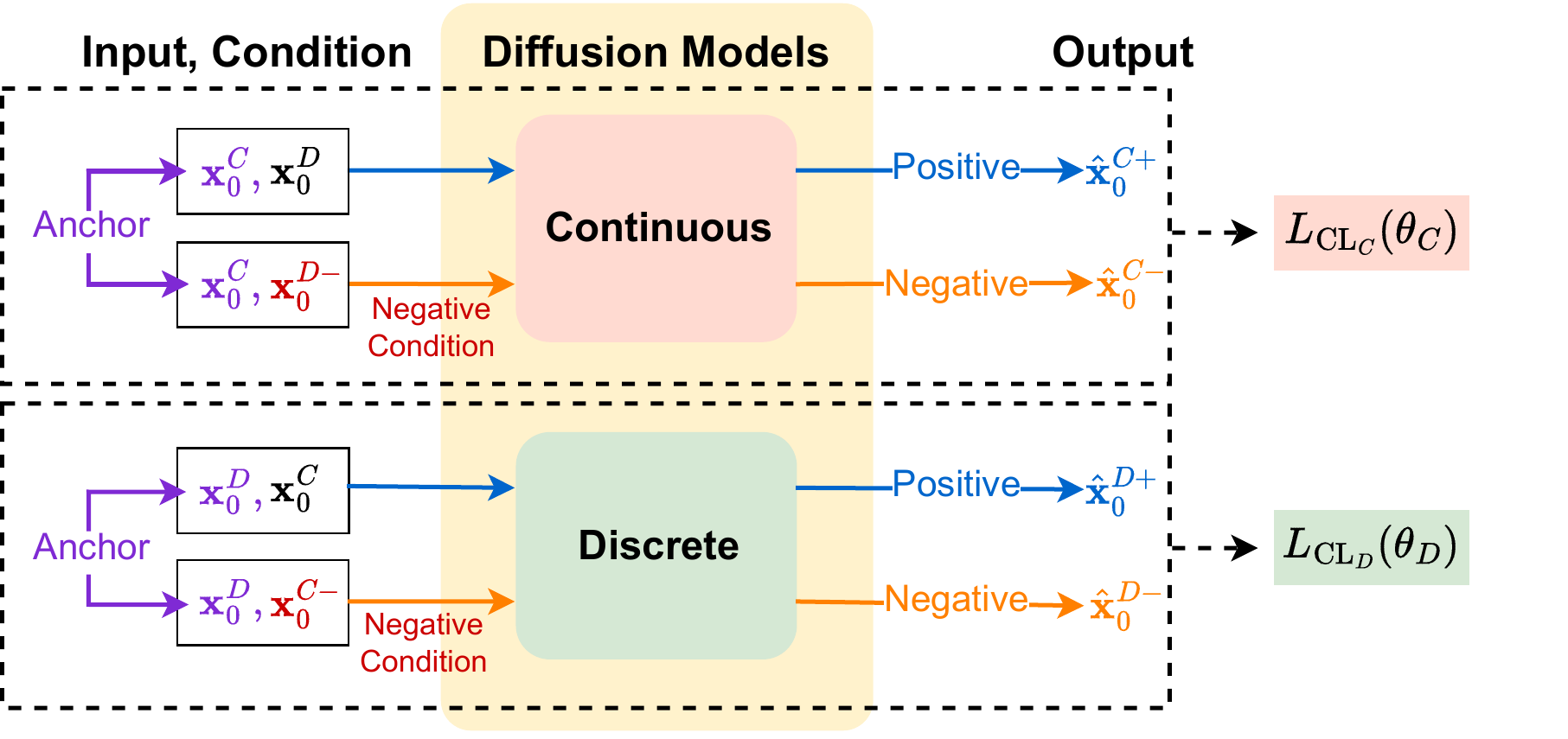}
        \caption{The proposed contrastive learning for tabular data. With a negative sampling method, we encourage the two diffusion models to generate samples that are closer to the positive samples and distinct from the negative samples.}
        \label{fig:method2}
        \vspace{-0.8em}
\end{figure}
\subsection{Contrastive Learning}\label{sec:contrastive}
To connect the two diffusion models further, we adopt a contrastive learning method for tabular data by utilizing the following triplet loss~\cite{triplet}, which prefers positive samples to be close to the anchor. The objective function is as follows: \begin{equation}\label{eq:triplet}
\resizebox{0.89\hsize}{!}{%
$L_{\mathrm{CL}}(A, P, N)= \sum_{i=0}^S\big[\max{\{d(A_i,P_i)-d(A_i,N_i)+m, 0\}}\big],$
}
\end{equation}
where $A$ is an anchor, $P$ is a positive sample, $N$ is a negative sample, $d$ is a distance metric, $m$ is a margin between the positive and negative samples, and $S$ is the number of samples.

Fig.~\ref{fig:method2} shows an overall process of the contrastive learning with the anchor, positive, and negative samples. Our contrastive learning process is applied to the continuous and discrete diffusion models separately. 
For simplicity but without loss of generality, we describe a process mainly for the continuous diffusion model, and the constrastive learning for the discrete diffusion model follows a similar process. In our method, we set a real sample $\mathbf{x}_0^C$ as the anchor, and a generated sample $\hat{\mathbf{x}}_0^{C+}$ conditioned on $\mathbf{x}_0^D$ as the positive sample. For the negative sample, we generate $\hat{\mathbf{x}}_0^{C-}$ with negative condition $\mathbf{x}_0^{D-}$, which is an inappropriate counterpart for $\mathbf{x}_0^C$.

Due to the computationally expensive nature of the diffusion models, generating a sample requires a long time, and for every training iteration, generating positive and negative samples for contrastive learning via $T$ steps may significantly delay training time. For this reason, we estimate the positive and negative samples using the model's output. 
To be specific, we can predict $\hat{\mathbf{x}}_0^{C+}$ and $\hat{\mathbf{x}}_0^{C-}$ using the continuous diffusion model with the following equations: \begin{equation}\label{eq:cl_ps}
\hat{\mathbf{x}}_0^{C+}=(\mathbf{x}_t^C-\sqrt{1-\bar{\alpha}_t}\boldsymbol{\epsilon}_{\theta_C}(\mathbf{x}_t^C, t\mid \mathbf{x}_t^D))/\sqrt{\bar{\alpha}_t},
\end{equation}
\begin{equation}\label{eq:cl_ns}
\hat{\mathbf{x}}_0^{C-}=(\mathbf{x}_t^C-\sqrt{1-\bar{\alpha}_t}\boldsymbol{\epsilon}_{\theta_C}(\mathbf{x}_t^C, t\mid \mathbf{x}_t^{D-}))/\sqrt{\bar{\alpha}_t}.
\end{equation}
Likewise, for the discrete diffusion model, we can directly estimate $\hat{\mathbf{x}}_0^{D+}$ and $\hat{\mathbf{x}}_0^{D-}$, using $p_{\theta_D}(\hat{\mathbf{x}}_0^{D+}| \mathbf{x}_t^D,\mathbf{x}_t^{C})$ and $p_{\theta_D}(\hat{\mathbf{x}}_0^{D-}| \mathbf{x}_t^D,\mathbf{x}_t^{C-})$, respectively.

After generating positive and negative samples, we calculate the contrastive learning losses with Eq.~\eqref{eq:triplet}. For continuous variables  $\mathbf{x}_0^C$ ($A$), $\hat{\mathbf{x}}_0^{C+}$ ($P$), and $\hat{\mathbf{x}}_0^{C-}$ ($N$), we use Euclidean distance as a metric $d$; for discrete variables $\mathbf{x}_0^D$ ($A$), $\hat{\mathbf{x}}_0^{D+}$ ($P$), and $\hat{\mathbf{x}}_0^{D-}$ ($N$), we use cross-entropy. Then, we combine the diffusion model losses and the contrastive learning losses as follows: \begin{equation}
L_{\mathrm{C}}(\theta_C) = L_{\mathrm{Diff_C}}(\theta_C) + \lambda_C L_{\mathrm{CL_C}}(\theta_C),
\end{equation}
\begin{equation}
L_{\mathrm{D}}(\theta_D) = L_{\mathrm{Diff_D}}(\theta_D) + \lambda_D L_{\mathrm{CL_D}}(\theta_D),
\end{equation}
where  $L_{\mathrm{CL}_C}(\theta_C)$ and  $L_{\mathrm{CL}_D}(\theta_D)$ are the contrastive learning losses, and $L_{\mathrm{C}}(\theta_C)$ and $L_{\mathrm{D}}(\theta_D)$ are final losses for the continuous and discrete diffusion models, respectively, $0<\lambda_C<1$, and $0<\lambda_D<1$.

\begin{figure}[t]
        \centering
        \includegraphics[width=0.98\linewidth]{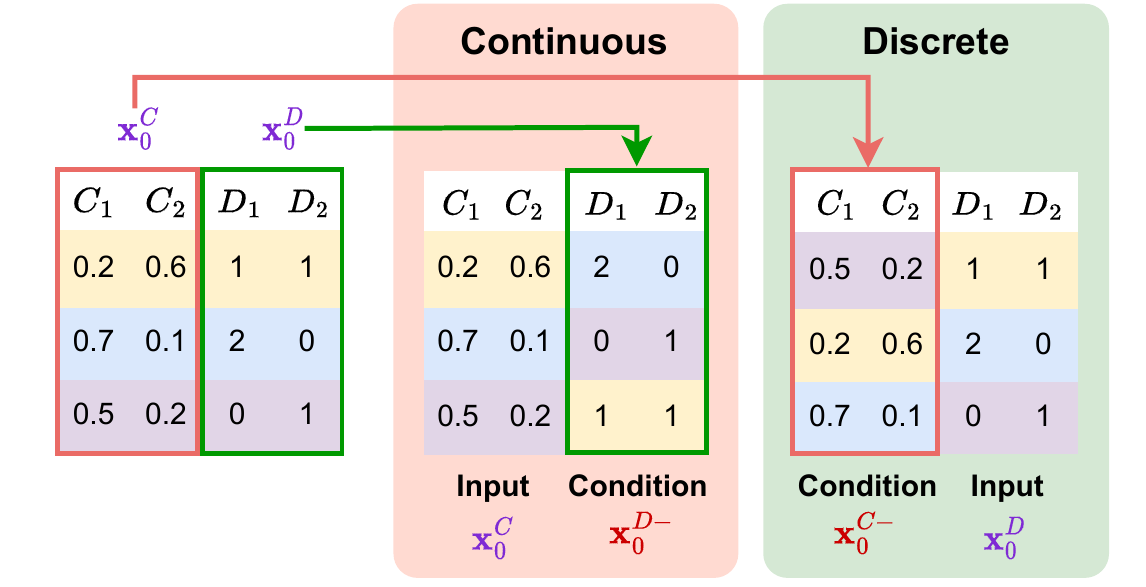}
        \caption{How to define the negative conditions. We randomly permute the continuous and discrete variable sets while maintaining their internal pairs and therefore, the inter-variable correlation does not make sense, i.e., they are not appropriate counterparts to each other.}
        \label{fig:negative_sampling}
\vspace{-1em}
\end{figure}

\noindent \textbf{Negative Condition}
We note that the negative conditions, $\mathbf{x}_0^{D-}$ and $\mathbf{x}_0^{C-}$, are the keys to generate the negative samples. To generate the negative samples, we focus on defining the negative conditions first. As shown in Fig.~\ref{fig:negative_sampling}, we make the negative conditions by randomly shuffling the continuous and discrete variable sets so that they do not match. For example, given $\mathbf{x}_0^C$ of the continuous diffusion model, its negative condition $\mathbf{x}_0^{D-}$ is from other random record's discrete part. The negative condition $\mathbf{x}_0^{C-}$ of $\mathbf{x}_0^D$ for the discrete diffusion model also follows the same method. In other words, the negative samples are generated by corrupting the inter-correlation between the continuous and discrete variable sets.
\begin{figure}[t]
\begin{algorithm}[H]
   \caption{Training}
   \label{alg:training}
\begin{algorithmic}
   \STATE Initialize $\theta_C$ and $\theta_D$
   \REPEAT
    \STATE $\mathbf{x}_0^C\sim q(\mathbf{x}_0^C)$, $\mathbf{x}_0^{D}\sim q(\mathbf{x}_0^{D})$, $t\sim \mathcal{U}(\{1,\dots,T\})$
     \STATE Compute $L_{\mathrm{Diff_C}}(\theta_C)$ and $L_{\mathrm{Diff_D}}(\theta_D)$
     \STATE Make negative conditions $\mathbf{x}_0^{C-}$ and $\mathbf{x}_0^{D-}$ 
     \STATE Generate positive samples $\hat{\mathbf{x}}_0^{C+}$ and $\hat{\mathbf{x}}_0^{D+}$ 
     \STATE Generate negative samples $\hat{\mathbf{x}}_0^{C-}$ and $\hat{\mathbf{x}}_0^{D-}$ 
     \STATE Compute $L_{\mathrm{CL}_C}(\theta_C)$ and $L_{\mathrm{CL}_D}(\theta_D)$
     \STATE $L_{\mathrm{C}}(\theta_C) \gets L_{\mathrm{Diff_C}}(\theta_C) + \lambda_C L_{\mathrm{CL_C}}(\theta_C)$
     \STATE $L_{\mathrm{D}}(\theta_D) \gets L_{\mathrm{Diff_D}}(\theta_D) + \lambda_D L_{\mathrm{CL_D}}(\theta_D)$
     \STATE Update $\theta_C$ and $\theta_D$
   \UNTIL{converged}
\end{algorithmic}
\end{algorithm}
\end{figure}

\begin{figure}[t]
\vspace{-2.5em}
\begin{algorithm}[H]
   \caption{Sampling}
   \label{alg:sampling}
\begin{algorithmic}
   \STATE $\hat{\mathbf{x}}_T^C\sim p(\mathbf{x}_T^C), \hat{\mathbf{x}}_T^{D}\sim p(\mathbf{x}_T^{D})$ 
   \FOR{$i=T, \dots, 1$}
    \STATE $\hat{\mathbf{x}}_{i-1}^C \sim p_{\theta_C}(\hat{\mathbf{x}}_{i-1}^C | \hat{\mathbf{x}}_{i}^C, \hat{\mathbf{x}}_{i}^D)$
    \STATE $\hat{\mathbf{x}}_{i-1}^D \sim p_{\theta_D}(\hat{\mathbf{x}}_{i-1}^D | \hat{\mathbf{x}}_{i}^D, \hat{\mathbf{x}}_{i}^C)$
   \ENDFOR
   \STATE {\bfseries return $\hat{\mathbf{x}}_0^C, \hat{\mathbf{x}}_0^D$}
\end{algorithmic}
\end{algorithm}
\vspace{-2em}
\end{figure}
\subsection{Training \& Sampling Algorithms}
Algorithm~\ref{alg:training} shows the overall training process of \texttt{CoDi}. We pre-process continuous variables to be $[-1, 1]$ using the min-max scaler, and use one-hot encoding for discrete variables. Then, we initialize two diffusion model parameters $\boldsymbol{\theta}_C$ and $\boldsymbol{\theta}_D$. We sample $t$ from Uniform distribution, and train the two conditional diffusion models with Eqs.~\eqref{eq:diff_con} and~\eqref{eq:diff_dis}, respectively. After that, we make negative conditions for contrastive learning with the method described in Section~\ref{sec:contrastive}. We generate positive and negative samples, and compute contrastive learning losses with Eq.~\eqref{eq:triplet}. Finally, we integrate the contrastive learning losses with the diffusion model losses, and update model parameters $\boldsymbol{\theta}_C$ and $\boldsymbol{\theta}_D$, respectively.

The detailed sampling process of our method is in Algorithm~\ref{alg:sampling}. Firstly, we sample each noisy vector from the corresponding prior distribution, and convert the noises into fake samples through $T$ steps. At this point, $\mathbf{x}_{i-1}^C$ and $\mathbf{x}_{i-1}^D$ are conditioned on both denoised samples in continuous and discrete models at the previous time step $i$. After sampling, we post-process the continuous and discrete outputs, using the reverse scaler and Argmax function, respectively. 

\begin{table*}[!ht]
\caption{Summarized experimental results in terms of sampling quality. We report averaged scores across the datasets. Full results are in Appendix~\ref{sec:apd_quality}. We highlight the best results in light blue, and the second best with underline. }
\label{table:samplingquality}
\begin{center}
\begin{sc}
\begin{tabular}{lccccccccccc}
        \toprule
        \multirow{2}{*}{Methods} & \multicolumn{3}{c}{Binary} &&  \multicolumn{3}{c}{Multi-class} &&  \multicolumn{3}{c}{Regression} \\ \cmidrule{2-4} \cmidrule{6-8} \cmidrule{10-12}
        & Binary F1 && AUROC && Macro F1 && AUROC && $R^2$ && RMSE \\ 
        \midrule
        \texttt{Identity} & 0.4154 & &0.8119 && 0.6514 & & 0.8230 && 0.6673 && 0.3593 \\ \midrule
        \texttt{MedGAN} & 0.1523 && 0.5464 && 0.1537 && 0.5015 && -inf && inf \\ 
        \texttt{VEEGAN} & 0.2591 && 0.5520  && 0.1206  && 0.5082 && -inf && inf \\ 
        \texttt{CTGAN} & 0.3432 && 0.6745 && 0.2355 && 0.5546 && -inf && inf \\ 
        \texttt{TVAE} & 0.3188 && 0.6867 && 0.2361 && 0.5974 && -inf && inf \\ 
        \texttt{TableGAN} & 0.4078 && 0.7480 && 0.2715 && 0.6072 && \underline{-0.0704} & & \underline{1.0015} \\ 
        \texttt{OCT-GAN} & 0.3814 && 0.7350 && 0.3314 && 0.6434 && -0.0868 && 1.0210 \\ 
        \texttt{RNODE} & 0.3208 && 0.6651 && 0.3692 && 0.7037 && -0.3037 && 1.1270 \\
        \texttt{STaSy} & \underline{0.4559} && \underline{0.7961} && \underline{0.6078} && \underline{0.7997} && -1.3200 && 1.2227 \\ 
        \midrule
        \rowcolor{cyan!15} \texttt{CoDi} & \textbf{0.4726} && \textbf{0.8106} && \textbf{0.6221} && \textbf{0.8026} && \textbf{0.4794} && \textbf{0.6477} \\  
        \bottomrule
\end{tabular}
\end{sc}
\end{center}
\end{table*}

\section{Experiments}\label{sec:experiments}
In this section, we introduce our experimental environments and results. We demonstrate the performance of our proposed method in terms of \textit{the generative learning trilemma}. Detailed settings and hyperparameters are in Appendix~\ref{sec:apd_env}.

\subsection{Experimental Setup}\label{sec:exp_env}
\subsubsection{Datasets \& Baselines}
In our experiments, we select 11 datasets considering the number of continuous and discrete variables, and utilize 8 generative models from GANs to score-based generative models. Detailed information on datasets and baselines is in Appendix~\ref{sec:apd_dataset} and~\ref{sec:apd_baseline}.

\subsubsection{Evaluation Methods}
We strictly follow evaluation methods described in~\citet{stasy}. To evaluate the sampling quality, we train various classification/regression models with fake data, validate with real training data, and test them with real test data, called ``TSTR''~\cite{esteban2017realvalued, Jordon2019PATEGANGS}. We use F1 and AUROC as evaluation metrics for classification, and $R^2$ and RMSE for regression. We evaluate the model using 5 different fake samples by finding the best hyperparameter sets for classification/regression models. We test the best models with 5 fake samples and report the average score and standard deviation. 
To measure the sampling diversity, we utilize coverage~\cite{coverage}. For the sampling time, we measure wall-clock time taken to generate 10K fake samples. We measure the diversity and sampling time 5 times with different fake samples, and report their mean and standard deviation.

\subsection{Experimental Results}\label{sec:exp_results}

\subsubsection{Sampling Quality}
In Table~\ref{table:samplingquality}, we summarize experimental results on the sampling quality of each tabular data synthesis method. The results are averaged scores across the datasets. As shown, in classification tasks, \texttt{MedGAN} and \texttt{VEEGAN} show quite inferior performance to the others. More advanced GAN-based methods, i.e., \texttt{TableGAN} and \texttt{OCT-GAN}, and flow-based method, i.e., \texttt{RNODE}, perform to some degree. On the other hand, \texttt{STaSy} and \texttt{CoDi} show beyond-comparison results, especially, \texttt{CoDi} outperforms \texttt{STaSy} in all cases. 

In the regression task, half of the baseline methods perform poorly, showing impractical results. Among the 9 methods, only \texttt{CoDi} shows positive $R^2$, which means \texttt{CoDi} is not only able to improve modeling in discrete variables but also generate high-quality continuous variables.

\begin{figure}[t]
        \centering
        \resizebox{0.7\hsize}{!}{%
        \begin{subfigure}{\includegraphics[width=0.57\columnwidth]{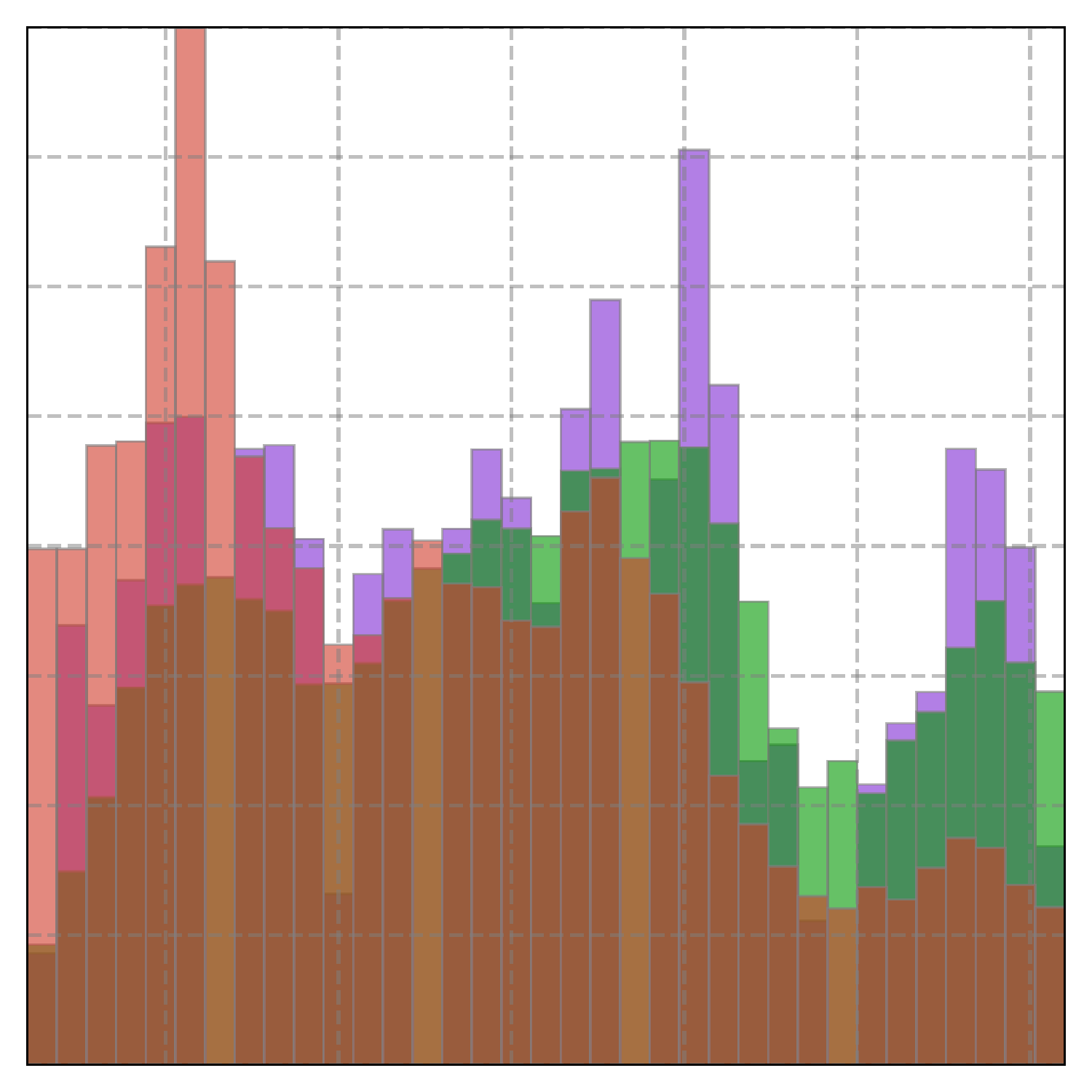}}
        \end{subfigure} \hfill
        \begin{subfigure}{\includegraphics[width=0.38\columnwidth]{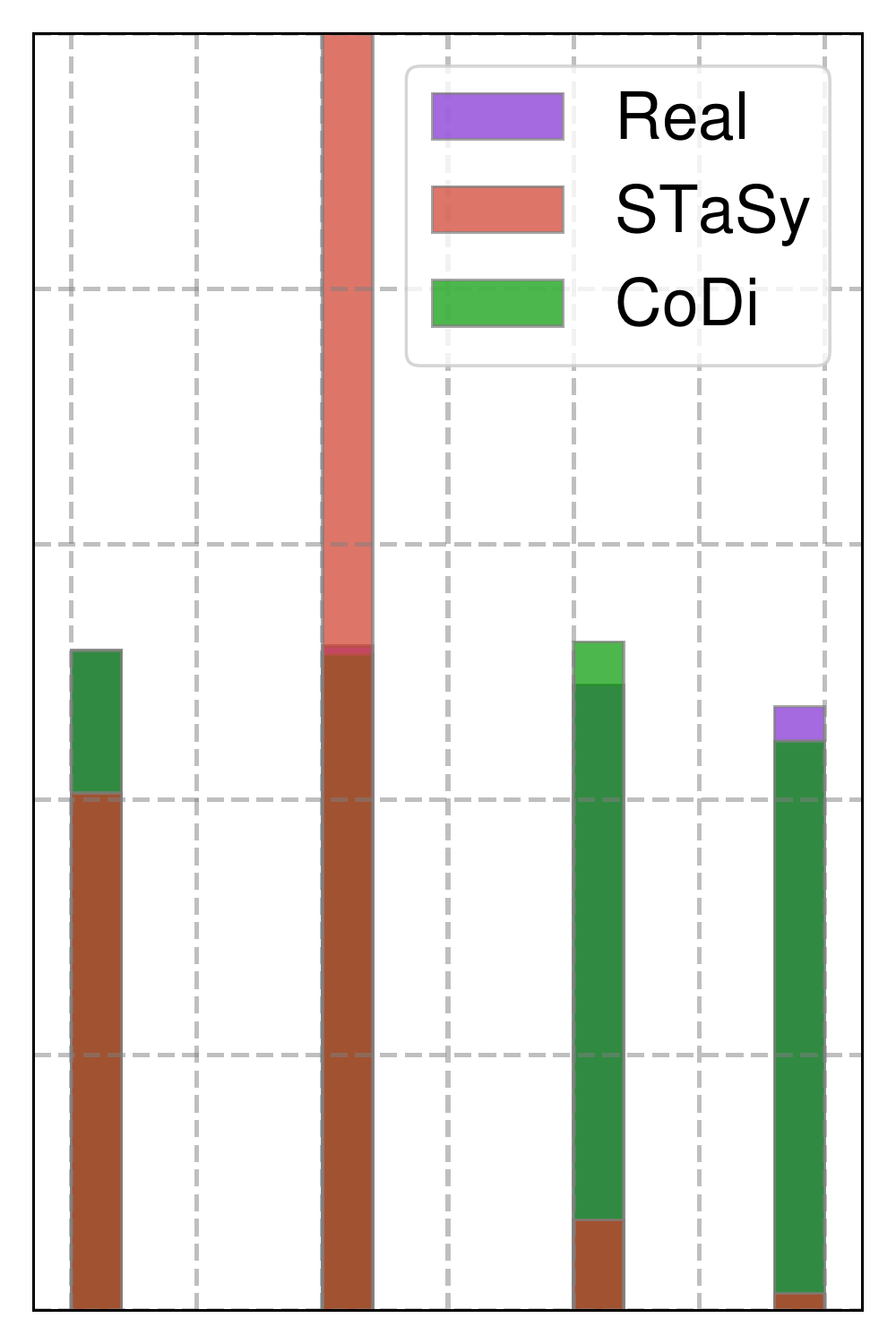}}
        \end{subfigure}}
        \caption{Comparison of column-wise distribution on real data, fake data by \texttt{STaSy} and \texttt{CoDi}. (Left) is for \textit{Days} (continuous) of \texttt{Bank} and (Right) is for \textit{Seasons} (discrete) of \texttt{Absent}.}
        \label{fig:histo}
\end{figure}

As shown in Fig.~\ref{fig:histo}, the fake data by \texttt{STaSy} hardly follows the real data distribution. Especially in Fig.~\ref{fig:histo} (Right), \texttt{STaSy} generates an abnormal number of samples that belongs to a specific category, i.e., the second category from the left, while the fake data by \texttt{CoDi} has a similar histogram to the real data.

\subsubsection{Sampling Diversity} 
To evaluate the diversity of the fake data, we use coverage, which is the ratio of fake samples that have at least one real sample within their $5^{th}$ nearest neighborhoods. Table~\ref{table:samplingdiversity} shows the averaged coverage scores across the datasets. Among the baselines, \texttt{MedGAN} and \texttt{VEEGAN} show poor performance, whereas \texttt{TableGAN} and \texttt{STaSy} show reliable diversity to some extent. However, \texttt{CoDi} exceeds other methods and particularly performs well in multi-class classification and regression datasets, as shown in Tables~\ref{tab:apd_diversity1} and~\ref{tab:apd_diversity2} of Appendix~\ref{sec:apd_diversity}. In \texttt{Stroke}, \texttt{TableGAN} and \texttt{CTGAN} show good performance in terms of the sampling quality, however, \texttt{CoDi} outperforms by large margins in terms of the diversity.

\begin{table}[t]
\begin{minipage}{0.46\linewidth}
\caption{Summarized sampling diversity results. Full results are in Appendix~\ref{sec:apd_diversity}. }
\setlength\tabcolsep{2pt}
\label{table:samplingdiversity}
\vskip 0.1in
\begin{center}
\begin{sc}
\begin{tabular}{lc}
\toprule
        Methods & Coverage  \\
        \midrule
        \texttt{MedGAN} & 0.0155 \\ 
        \texttt{VEEGAN} & 0.0019 \\ 
        \texttt{CTGAN} & 0.3834 \\ 
        \texttt{TVAE} & 0.3903 \\ 
        \texttt{TableGAN} & 0.5759 \\ 
        \texttt{OCT-GAN} & 0.2547 \\ 
        \texttt{RNODE} & 0.3841 \\ 
        \texttt{STaSy} & \underline{0.5771} \\ 
        \midrule
        \texttt{CoDi} & \cellcolor{cyan!15} \textbf{0.6931} \\ 
\bottomrule
\end{tabular}
\end{sc}
\end{center}
\end{minipage}
 \hfill
 \begin{minipage}{0.46\linewidth}
\caption{Summarized sampling time results. Full results are in Appendix~\ref{sec:apd_time}.}
\setlength\tabcolsep{2pt}
\label{table:samplingtime}
\begin{center}
\begin{sc}
\begin{tabular}{lc}
\toprule
Methods & Runtime  \\
\midrule
\texttt{MedGAN} & 0.0200 \\
\texttt{VEEGAN}  & \underline{0.0169} \\
\texttt{CTGAN} & 0.1260 \\
\texttt{TVAE}&\cellcolor{cyan!15} \textbf{0.0140} \\
\texttt{TableGAN}   & 0.0224 \\
\texttt{OCT-GAN}   & 0.6008 \\
\texttt{RNODE} & 103.1449 \\
\texttt{STaSy} & 4.6417 \\
\midrule
\texttt{CoDi}   & 0.5187 \\
\bottomrule
\end{tabular}
\end{sc}
\end{center}
\end{minipage}
\end{table}

\begin{table}[t]
\caption{Experimental results on discrete data trained in continuous and discrete spaces. We report the F1 and coverage, and highlight the best results in bold.}
\label{table:ablation_space}
\begin{center}
\setlength\tabcolsep{1pt}

\begin{small}
\begin{sc}
\begin{tabular}{lccccc}

\toprule

\multirow{2}{*}{Datasets} & \multicolumn{2}{c}{Continuous space} && \multicolumn{2}{c}{Discrete space} \\
\cmidrule{2-3} \cmidrule{5-6}
& F1 score & Coverage & & F1 score & Coverage\\
\midrule
        \texttt{Car} & 0.891\scriptsize{±0.021} & 0.639\scriptsize{±0.013} && \textbf{0.949\scriptsize{±0.021}} & \textbf{0.684\scriptsize{±0.004}} \\ 
        \texttt{Clave} & 0.591\scriptsize{±0.066} & 0.695\scriptsize{±0.007} &&\textbf{ 0.624\scriptsize{±0.073}} & \textbf{0.764\scriptsize{±0.006} }\\ 
        \texttt{Nursery} & 0.780\scriptsize{±0.016} & 0.553\scriptsize{±0.002} && \textbf{0.823\scriptsize{±0.036} }& \textbf{0.568\scriptsize{±0.003}} \\ 
        \texttt{Phishing} & 0.915\scriptsize{±0.008} & 0.127\scriptsize{±0.003} && \textbf{0.931\scriptsize{±0.012} }& \textbf{0.644\scriptsize{±0.008}} \\ 
\bottomrule
\end{tabular}
\end{sc}
\end{small}
\end{center}
\vspace{-1em}
\end{table}

\subsubsection{Sampling Time}
In Table~\ref{table:samplingtime}, we summarize sampling time. \texttt{MedGAN} and \texttt{VEEGAN} take short runtime, but show poor performance in terms of the sampling quality and diversity. Both \texttt{TVAE} and \texttt{TableGAN} also show fast speed, but the former generates less diverse samples, and the fake data by the latter shows low quality, which indicates they are not well-balanced in terms of \textit{the generative learning trilemma}. However, \texttt{STaSy} and \texttt{CoDi} can sample in reliable runtime with high quality and diversity. Especially, \texttt{CoDi} shows about 9 times faster speed than that of \texttt{STaSy}, and is comparable to other GAN-based methods in terms of the sampling time.

In \texttt{CoDi}, the input dimensions of the models are decreased by using two diffusion models for the two variable types. Moreover, our network design for each diffusion model allows for reducing the learnable parameters by adapting the U-Net-based skip connection instead of residual blocks. 
As a result, the model's complexity and training difficulty can be reduced considerably, and \texttt{CoDi} is able to balance \textit{the generative learning trilemma} more.



\subsection{Continuous vs. Discrete Spaces for Discrete Data}\label{sec:space}
We claim that discrete variables should be treated in discrete spaces, unlike the existing methods that process them in continuous spaces along with continuous variables. To verify the claim, we examine which is the appropriate space for discrete variables by training them in continuous and discrete spaces. We introduce the following 4 datasets for experiments: \texttt{Car}, \texttt{Clave}, \texttt{Nursery}, and \texttt{Phishing}, which only contain discrete variables. More information is in Appendix~\ref{sec:apd_dataset}. The diffusion models for this experiment are the continuous and discrete diffusion models of \texttt{CoDi}, where the conditioning and contrastive learning are omitted. 

\begin{figure}[t]
        \centering
        \includegraphics[width=0.85\columnwidth]{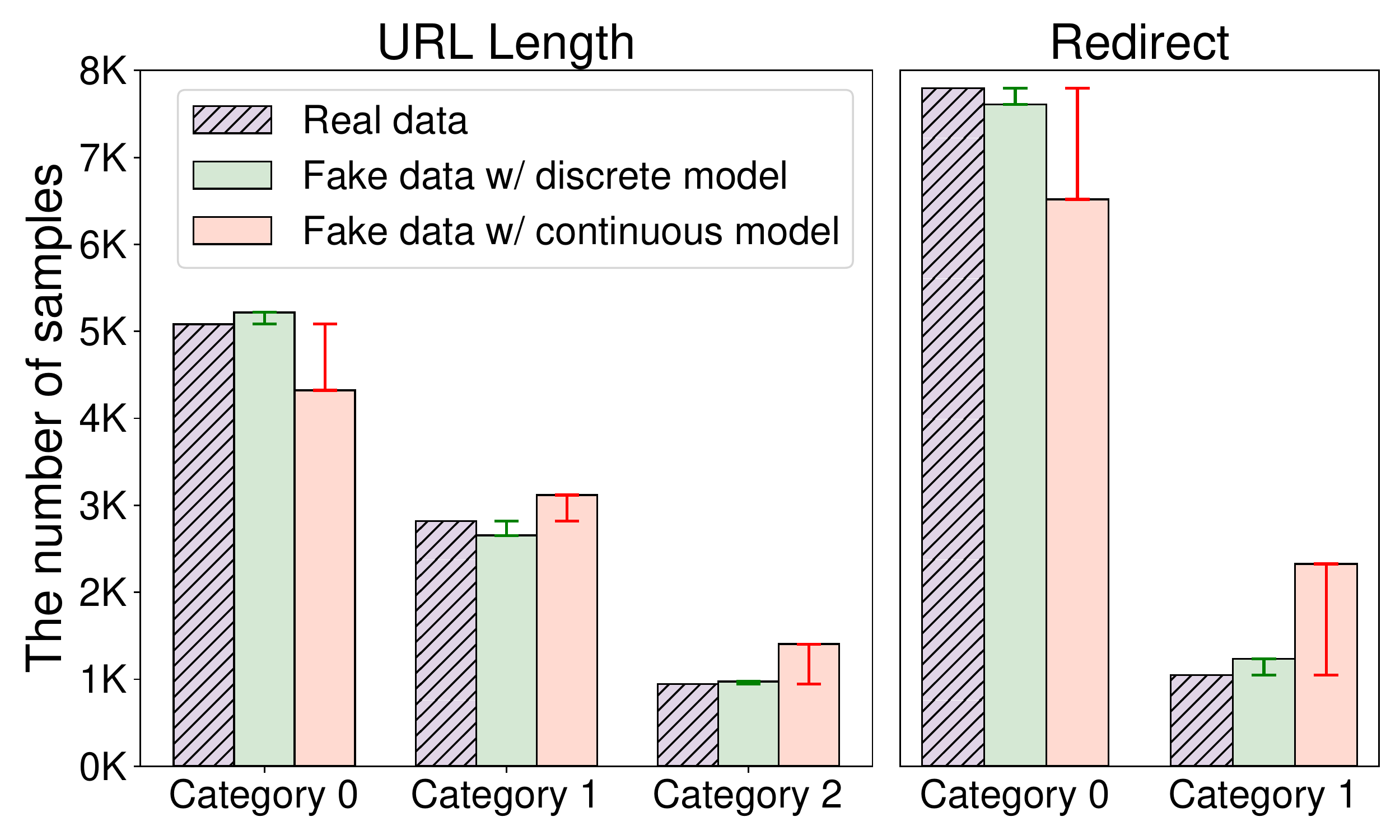}
        \caption{The number of samples for each category in discrete columns of \texttt{Phishing}. We selectively compare 2 columns (Left) \textit{URL Length}, which has 3 categories, and (Right) \textit{Redirect}, which is a binary class column. Green and red lines represent the gap between the real number of samples in each category and that of fake data. }
        \label{fig:discrete_v}
\end{figure}

As shown in Table~\ref{table:ablation_space}, in all cases, the F1 and coverage scores of the fake data by the discrete diffision model are better than those of the continuous diffusion model.
In Fig.~\ref{fig:discrete_v}, we compare the number of samples for each category of discrete columns between real data and fake data generated by the continuous and discrete diffusion models. As shown in Fig.~\ref{fig:discrete_v} (Left), the continuous diffusion model fails to retain the real number of each category. Especially in Category 1 of Fig.~\ref{fig:discrete_v} (Right), fake data by the continuous diffusion model generates more samples than the real data. The results show that the space where mixed-type variables are handled significantly affects the generation performance.

\begin{figure}[t] 
        \centering
        \includegraphics[trim={0cm 0cm 0cm 0cm},clip, width=1\linewidth]{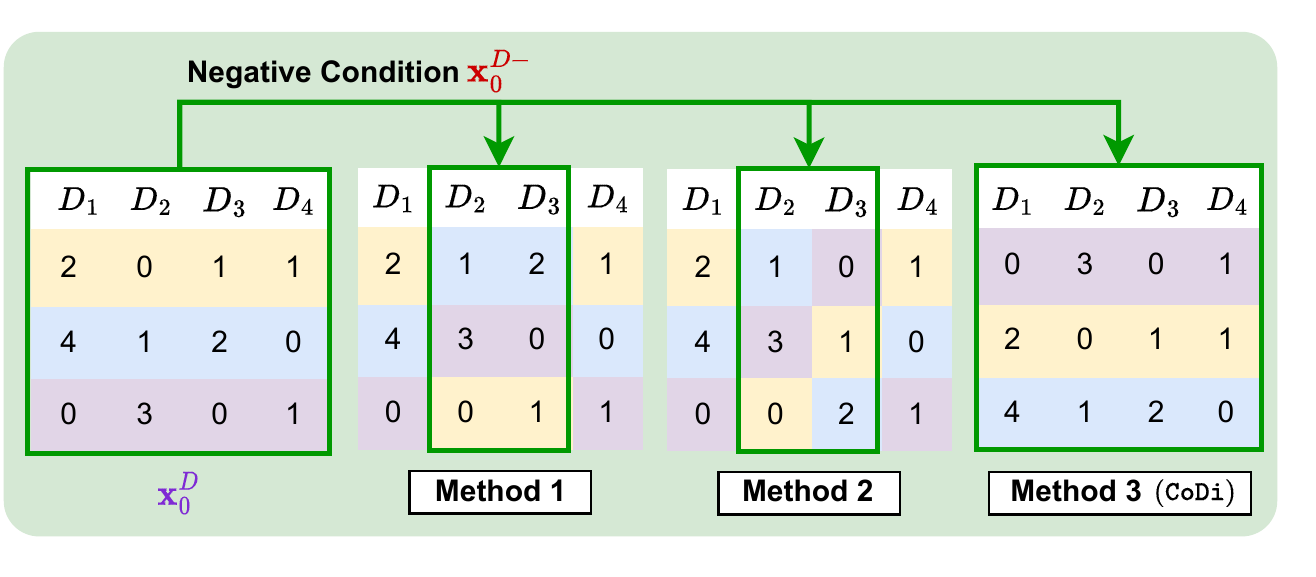}
        \caption{How to define negative conditions. \textit{Method 1} and \textit{Method 2} corrupt the inter-correlations between discrete variables, whereas \textit{Method 3} maintains them.
        }
        \label{fig:negative_condition}
\end{figure}


\begin{table}[t]
\caption{Experimental results on 3 negative sampling methods. We report F1 for classification data, $R^2$ for regression data, and coverage. The best results are highlighted in bold.}
\setlength\tabcolsep{1.5pt}
\label{table:ablation_negativesampling}
\vskip 0.1in
\begin{center}
\begin{small}
\begin{sc}
\begin{tabular}{llccc}
\toprule
Methods & Metrics & \texttt{Heart} &\texttt{Faults} &\texttt{Insurance} \\
\midrule
\multirow{2}{*}{Method 1} & F1 ($R^2$) & 0.857\scriptsize{±0.048} & 0.699\scriptsize{±0.041} & (0.567\scriptsize{±0.343})\\
 & Coverage & 0.847\scriptsize{±0.028} & 0.191\scriptsize{±0.012} &0.128\scriptsize{±0.020}\\
 \midrule
\multirow{2}{*}{Method 2} & F1 ($R^2$) & 0.850\scriptsize{±0.039} & 0.709\scriptsize{±0.042} & (0.554\scriptsize{±0.402})\\
 & Coverage & 0.820\scriptsize{±0.022} & 0.247\scriptsize{±0.005} &0.189\scriptsize{±0.024}\\
 \midrule
 {Method 3} & F1 ($R^2$) &\textbf{0.872\scriptsize{±0.039}} &\textbf{0.715\scriptsize{±0.046}} & (\textbf{0.575\scriptsize{±0.398}})\\
(\texttt{CoDi}) & Coverage &\textbf{0.949\scriptsize{±0.012}} &\textbf{0.270\scriptsize{±0.017}} &\textbf{0.262\scriptsize{±0.020}}\\
\bottomrule
\end{tabular}
\end{sc}
\end{small}
\end{center}
\vskip -0.1in
\end{table}

\subsection{Negative Sampling Methods}
Since the contrastive learning for tabular data has rarely been studied, in this section, we provide a comparison of possible negative sampling methods in terms of the sampling quality and diversity. We define 3 methods for the negative condition, as shown in Fig.~\ref{fig:negative_condition}. Let us define the discrete condition for the continuous diffusion model first for convenience. Firstly, for \textit{Method 1} and \textit{Method 2}, we randomly select two columns from discrete variables for the negative condition. Then, we shuffle the rows of the two columns, for \textit{Method 1}, retaining the pair of the two columns, and for \textit{Method 2}, without maintaining the pair of columns.
For \textit{Method 3}, which is our proposed method, we shuffle the rows of discrete variables altogether, maintaining the pair. The continuous condition can be defined by following the same process. 

In Table~\ref{table:ablation_negativesampling}, we compare the negative sampling methods. Although all negative sampling methods show reasonable performance in our experiment with 3 datasets, \textit{Method 3} shows the best performance in terms of the sampling quality and diversity. The first two methods disrupt not only the relationship between continuous and discrete variables but also the inter-correlations between each variable type, increasing the training difficulty of the model. On the other hand, \textit{Method 3} retains the latter, which can result in the outcome.

\begin{figure}[t]
        \begin{subfigure}\centering{\includegraphics[width=0.31\linewidth]{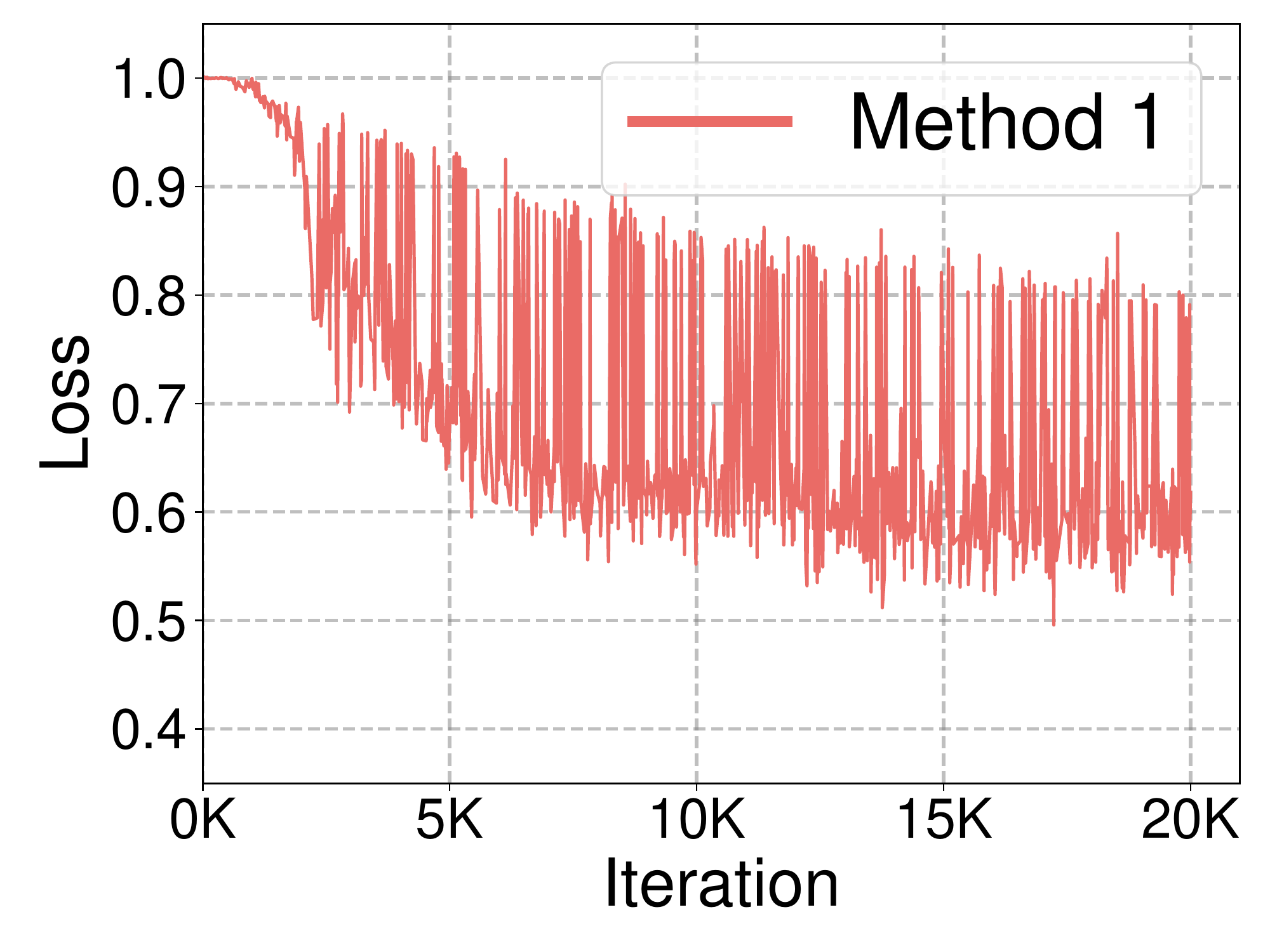}}
        \end{subfigure} 
        \begin{subfigure}\centering{\includegraphics[width=0.31\linewidth]{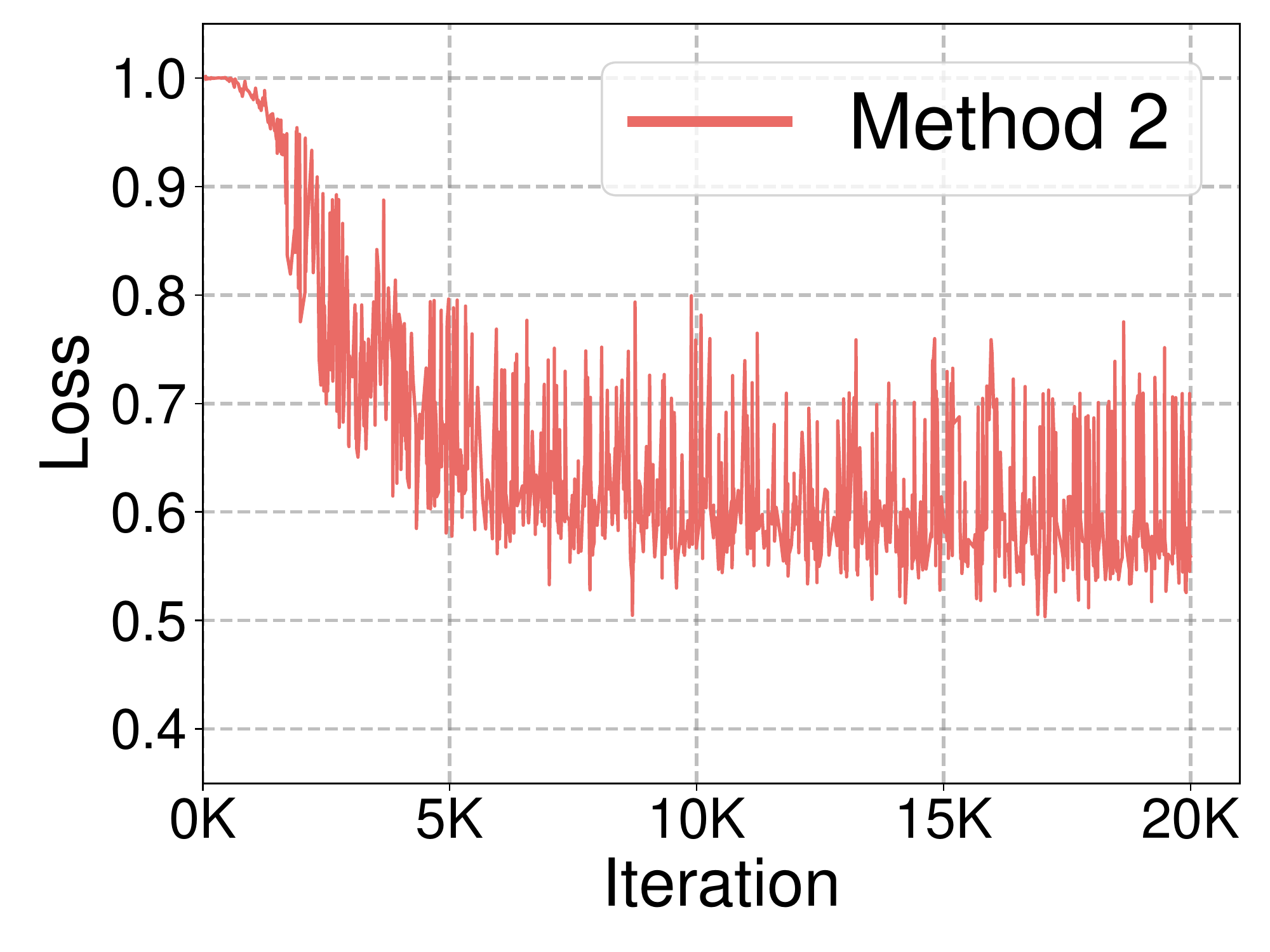}}
        \end{subfigure} 
        \begin{subfigure}\centering{\includegraphics[width=0.31\linewidth]{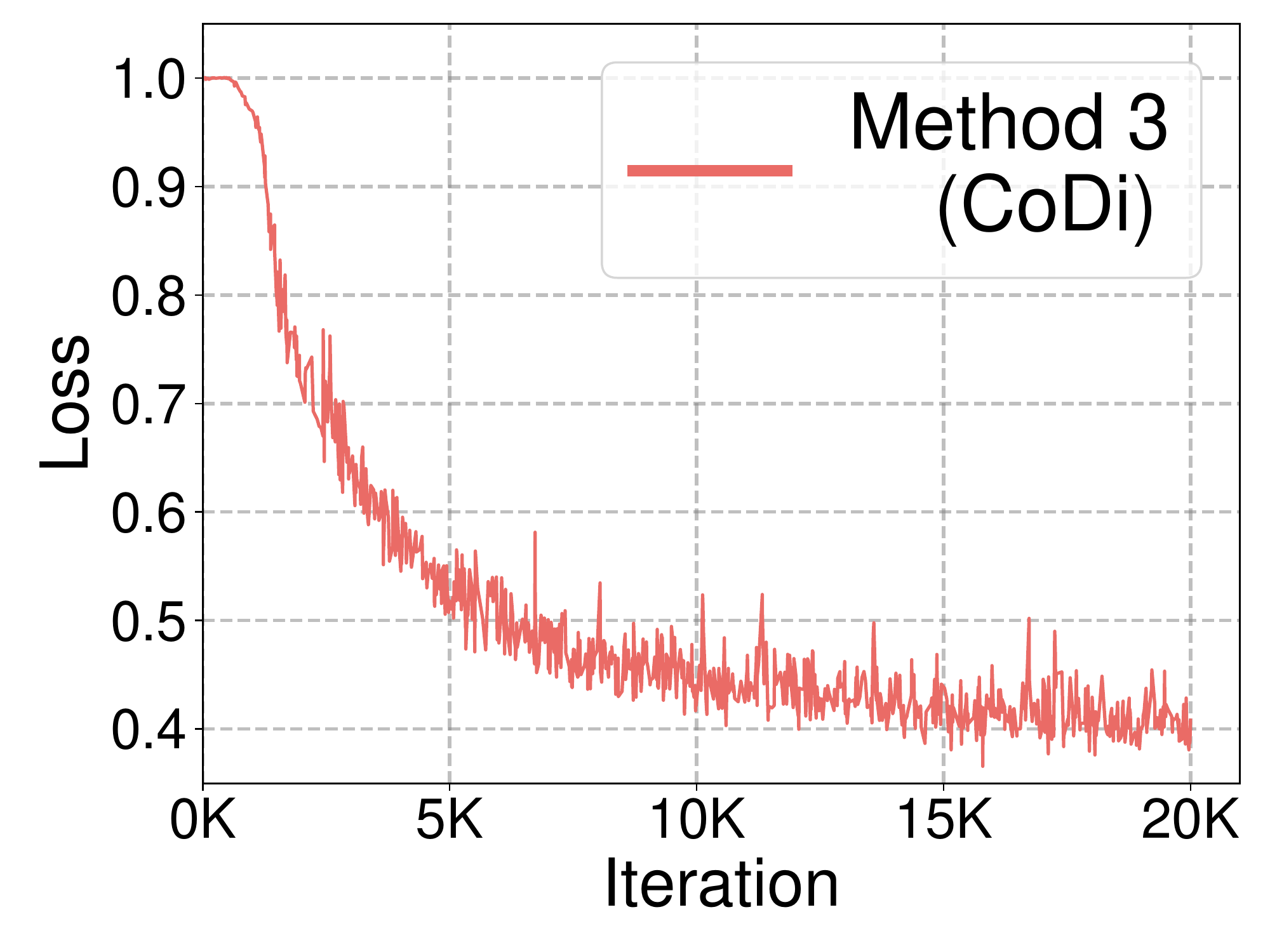}}
        \end{subfigure} \newline
        \begin{subfigure}\centering{\includegraphics[width=0.31\linewidth]{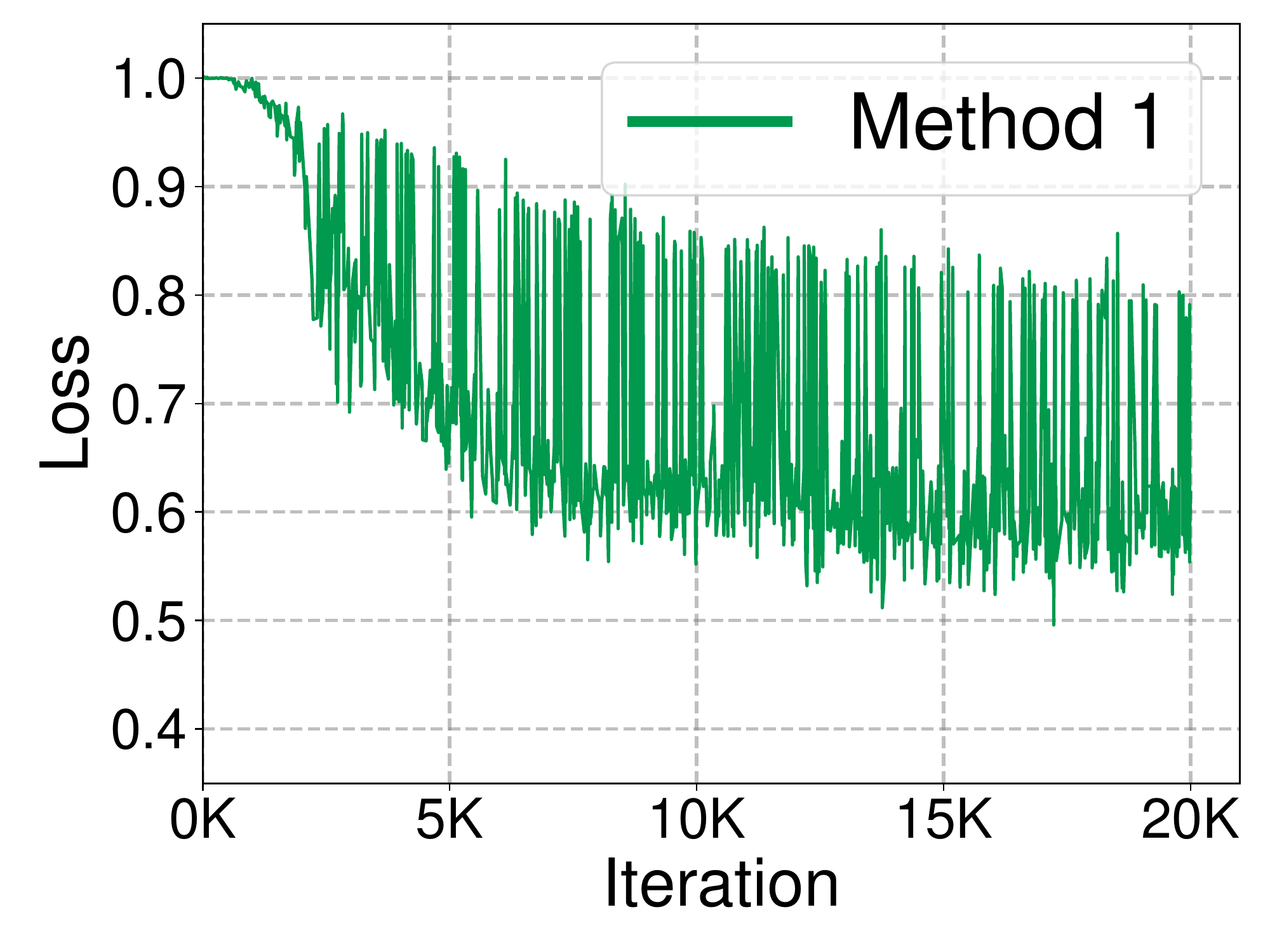}}
        \end{subfigure} 
        \begin{subfigure}\centering{\includegraphics[width=0.31\linewidth]{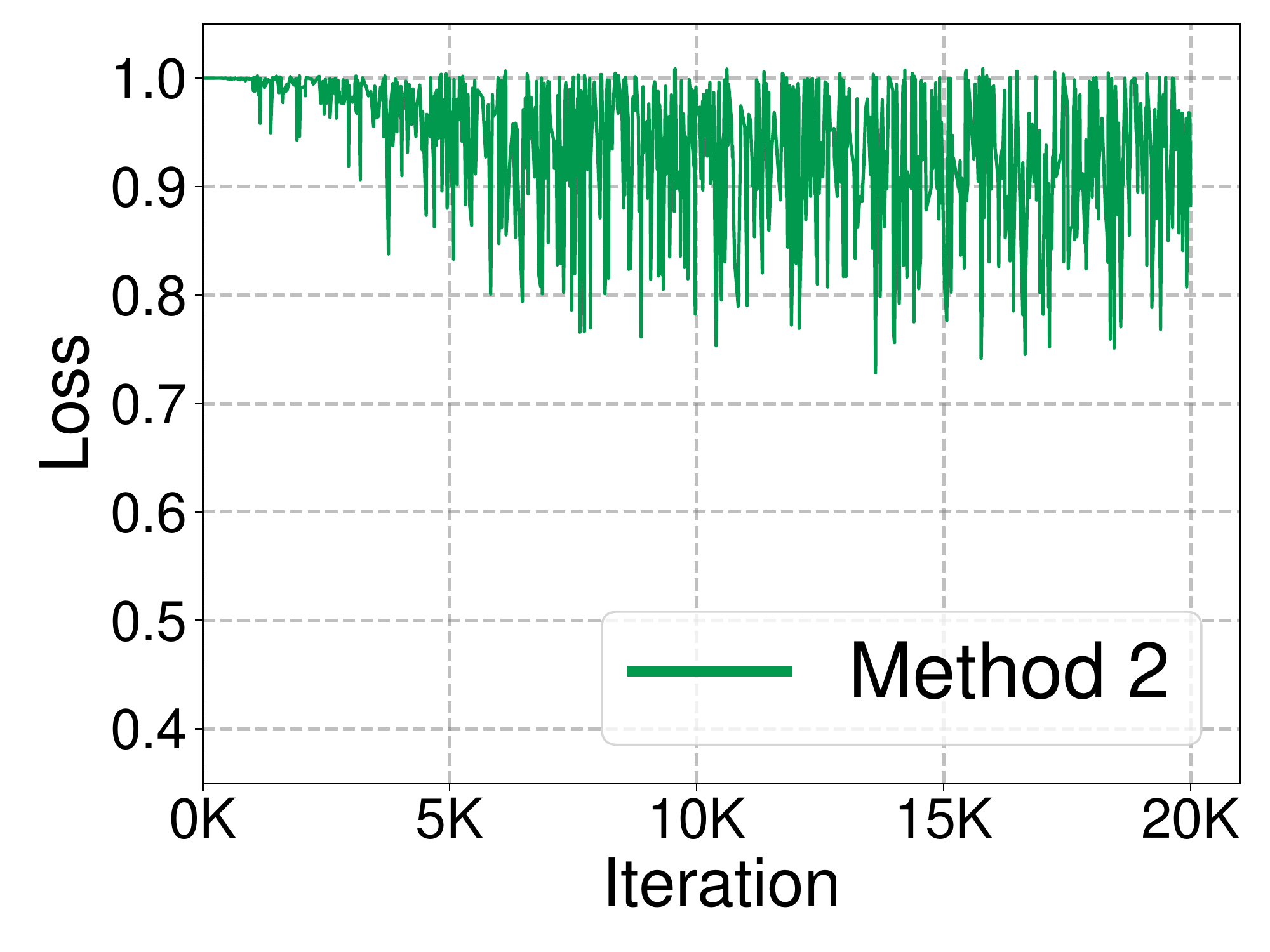}}
        \end{subfigure} 
        \begin{subfigure}\centering{\includegraphics[width=0.31\linewidth]{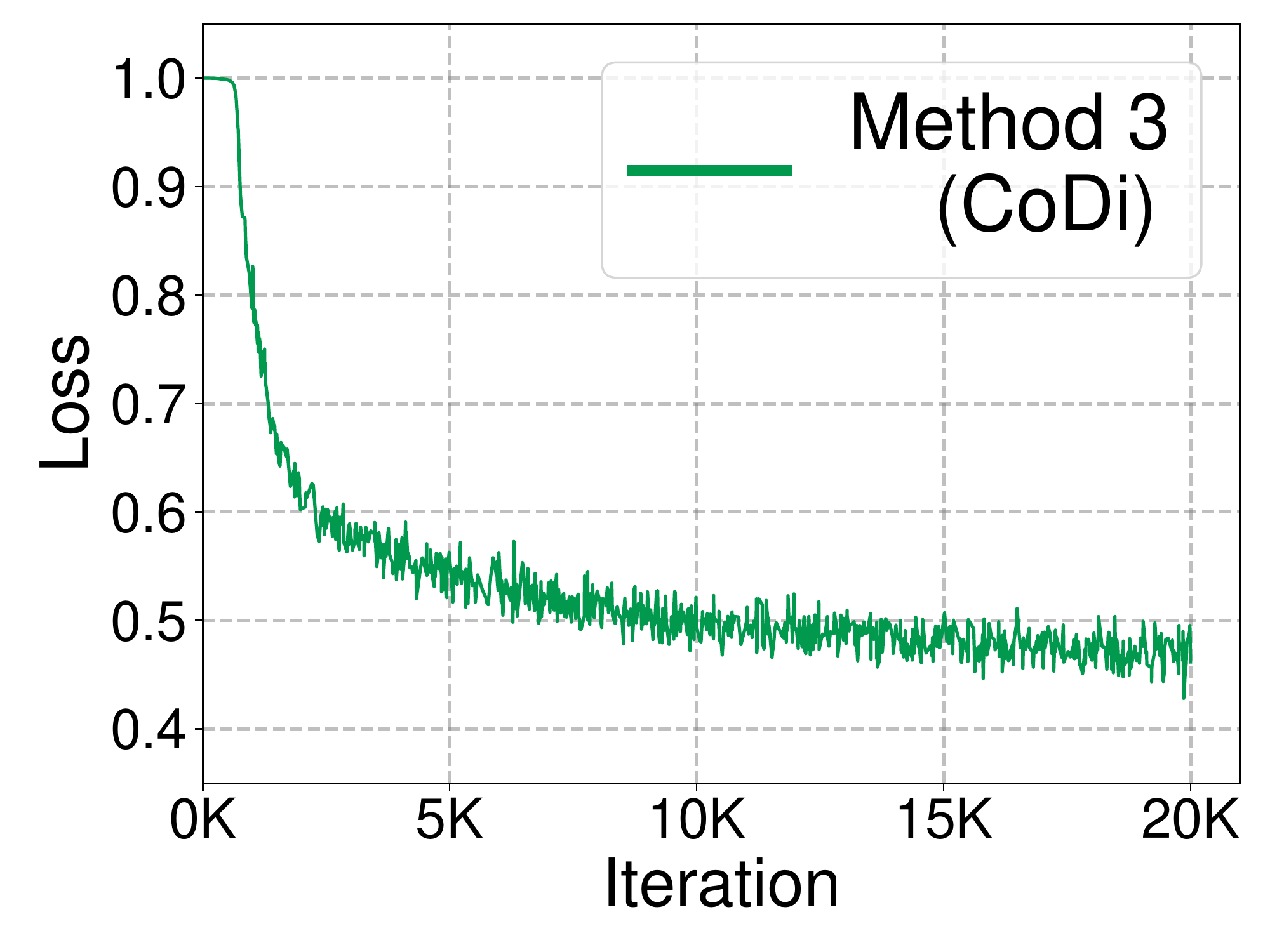}}
        \end{subfigure} 
        \caption{The training loss curves of triplet loss with respect to the negative sampling method in \texttt{Faults}. Red and green lines mean $L_{\mathrm{CL_C}}(\theta_C)$ and $L_{\mathrm{CL_D}}(\theta_D)$, respectively. (Left) are for \textit{Method 1}, (Middle) are for \textit{Method 2}, and (Right) are for \textit{Method 3}.}
        \label{fig:ns_loss}
\end{figure}

Fig.~\ref{fig:ns_loss} presents the training loss curves of the contrastive learning. 
As shown, the loss curves of the first two methods fluctuate drastically, while the loss curves of \textit{Method 3} converges stably, 
which represents the easiness of training.


\begin{table}[t]
\caption{Ablation study on the efficacy of the contrastive learning. We report F1 for classification data, $R^2$ for regression data, and coverage. 
}
\label{table:ablation_contrastive}
\vskip 0.1in
\begin{center}
\setlength\tabcolsep{0.3pt}
\begin{small}
\begin{sc}
\begin{tabular}{lccccc}
\toprule
\multirow{2}{*}{Datasets} & \multicolumn{2}{c}{\texttt{CoDi} w/o CL} && \multicolumn{2}{c}{\texttt{CoDi}}\\
\cmidrule{2-3} \cmidrule{5-6}
 & F1 ($R^2$) & Coverage && F1 ($R^2$) & Coverage\\
\midrule
        \texttt{Bank} & 0.527\scriptsize{±0.032} & \textbf{0.699\scriptsize{±0.003}} && \textbf{0.566\scriptsize{±0.014}}& 0.687\scriptsize{±0.002} \\ 
        \texttt{Heart} & \textbf{0.886\scriptsize{±0.043}} & 0.879\scriptsize{±0.017} && 0.872\scriptsize{±0.039}& \textbf{0.949\scriptsize{±0.012}} \\ 
        \texttt{Seismic} & 0.210\scriptsize{±0.064} & \textbf{0.380\scriptsize{±0.016}} && \textbf{0.305\scriptsize{±0.040}} & 0.359\scriptsize{±0.005} \\ 
        \texttt{Stroke} & 0.129\scriptsize{±0.036} & 0.651\scriptsize{±0.020} && \textbf{0.147\scriptsize{±0.016}} & \textbf{0.919\scriptsize{±0.008}} \\ 
        \midrule
        \texttt{CMC} & 0.484\scriptsize{±0.024} & 0.932\scriptsize{±0.011} && \textbf{0.503\scriptsize{±0.008}} & \textbf{0.934\scriptsize{±0.015}} \\ 
        \texttt{Customer} & 0.350\scriptsize{±0.008} & 0.789\scriptsize{±0.019} && \textbf{0.352\scriptsize{±0.015}}& \textbf{0.833\scriptsize{±0.021}}\\ 
        \texttt{Faults} & 0.705\scriptsize{±0.047} & \textbf{0.272\scriptsize{±0.016}} && \textbf{0.715\scriptsize{±0.046}} & 0.270\scriptsize{±0.017} \\ 
        \texttt{Obesity} & 0.912\scriptsize{±0.038} & \textbf{0.777\scriptsize{±0.018}} && \textbf{0.919\scriptsize{±0.034}}& 0.742\scriptsize{±0.015}\\ 
        \midrule
        \texttt{Absent} & (-0.026\scriptsize{±0.036}) & 0.801\scriptsize{±0.009} && (\textbf{0.095\scriptsize{±0.022}}) & \textbf{0.843\scriptsize{±0.023}} \\ 
        \texttt{Drug} & (0.748\scriptsize{±0.074}) & 0.813\scriptsize{±0.013} && (\textbf{0.768\scriptsize{±0.049}}) & \textbf{0.827\scriptsize{±0.046}} \\ 
        \texttt{Insurance} & (0.531\scriptsize{±0.308}) & 0.218\scriptsize{±0.028} && (\textbf{0.575\scriptsize{±0.398}}) & \textbf{0.262\scriptsize{±0.020}} \\
\bottomrule
\end{tabular}
\end{sc}
\end{small}
\end{center}
\vskip -0.1in
\end{table}

\subsection{Ablation Study on Contrastive Learning}
As shown in Table~\ref{table:ablation_contrastive}, we conduct ablation experiments to show the efficacy of the contrastive learning on \texttt{CoDi}. `\texttt{CoDi} w/o CL' means the model trained without contrastive learning. In \texttt{Bank}, \texttt{Seismic}, \texttt{Faults}, and \texttt{Obesity}, the contrastive learning improves F1 scores, and in \texttt{Heart}, it significantly enhances the sampling diversity. In other datasets, it enhances not only the sampling quality but also diversity. 
The results demonstrate that the design choice of the anchor, positive, and negative samples is reasonable and the contrastive learning is properly performed to synthesize a better sample.

\section{Conclusions}
Synthesizing realistic tabular data is one of the utmost tasks as tabular data is a frequently used data format. However, modeling tabular data is of non-trivial due to the nature of tabular data, which consists of mixed data types. To this end, we introduce the set of diffusion models for continuous and discrete variables. By conditioning the diffusion models at every training iteration using each other's outputs, the models are able to co-evolve maintaining the correlation of continuous and discrete variables. Moreover, the binding between two models is further reinforced by bringing the contrastive learning to our training. 

In our experiments with 11 real-world benchmark datasets and 8 baselines, \texttt{CoDi} outperforms others in terms of sampling quality and diversity. Compared to \texttt{STaSy}, the recent state-of-the-art tabular data synthesis method, our method improves the sampling time by about 90\%, owing to the reduced training difficulty. Consequently, \texttt{CoDi} achieves well-balanced \textit{generative learning trilemma}, representing remarkable advancements in tabular data synthesis.

\noindent \textbf{Limitations} \texttt{CoDi} is designed for better learning of continuous and discrete variables simultaneously, which means that our method may not be applicable when there are only continuous or discrete variables.
However, real-world tabular data typically has mixed types.

\noindent \textbf{Societal impacts} As the generated fake data become realistic, there is a risk of being abused by people trying to achieve their immoral purposes. This leads to the research topic of protecting privacy in generative models being actively studied~\cite{itgan, hyeong2022empirical}. 

\section*{Acknowledgements}
Chaejeong Lee and Jayoung Kim equally contributed. Noseong Park is the corresponding author. This work was supported by the Institute of Information \& communications Technology Planning \& Evaluation (IITP) grant funded by the Korea government (MSIT) (90\% from No. 2021-0-00231, Development of Approximate DBMS Query Technology to Facilitate Fast Query Processing for Exploratory Data Analysis and 10\% from No. 2020-0-01361, Artificial Intelligence Graduate School Program (Yonsei University)).

\medskip


\bibliography{example_paper}
\bibliographystyle{icml2023}

\newpage
\appendix
\onecolumn
\section{Preliminary Experiment}\label{sec:apd_toy}

\begin{figure}[h]
        \centering
        \begin{subfigure}{\includegraphics[width=0.25\columnwidth]{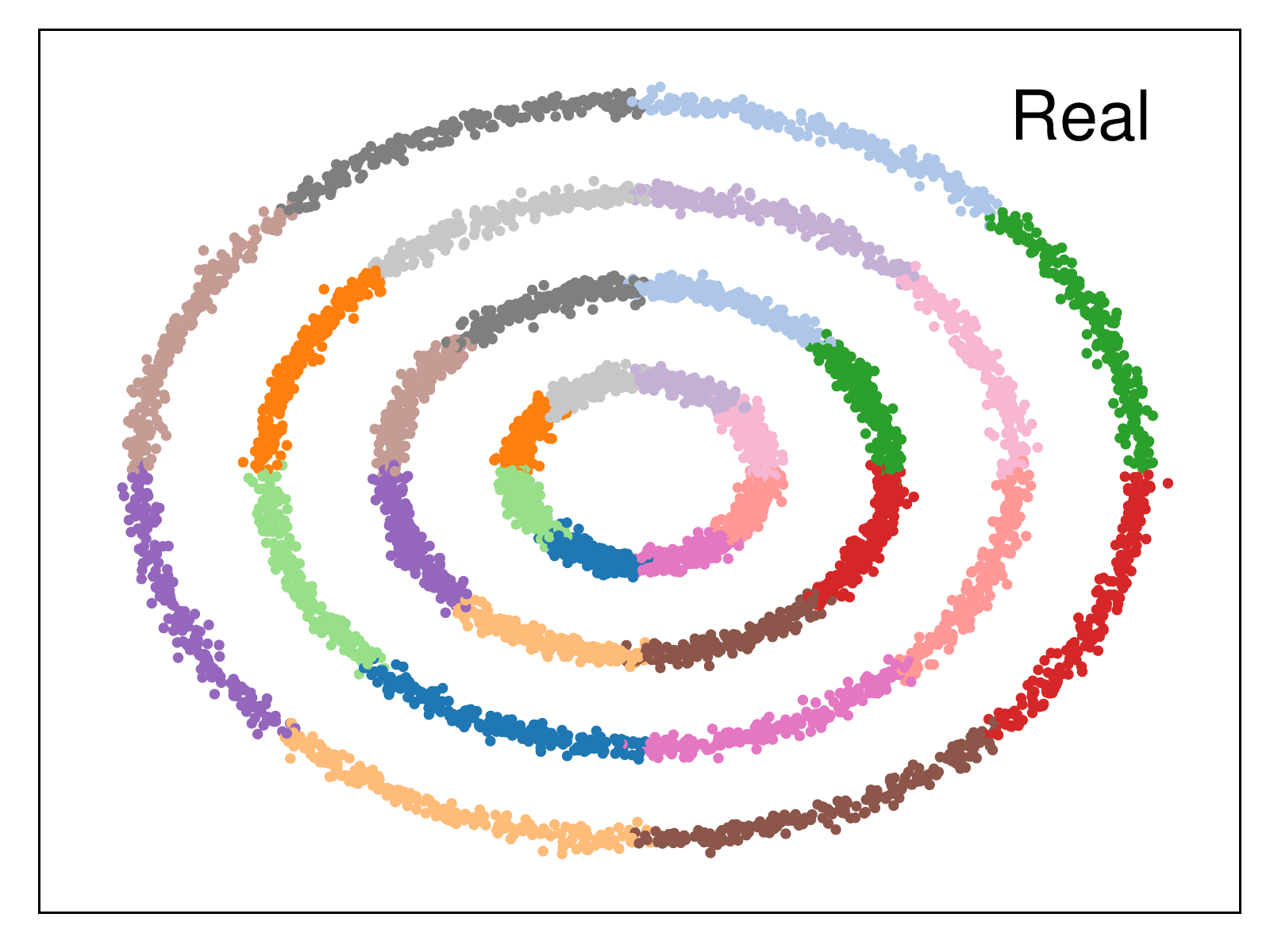}}
        \end{subfigure}
        \begin{subfigure}{\includegraphics[width=0.25\columnwidth]{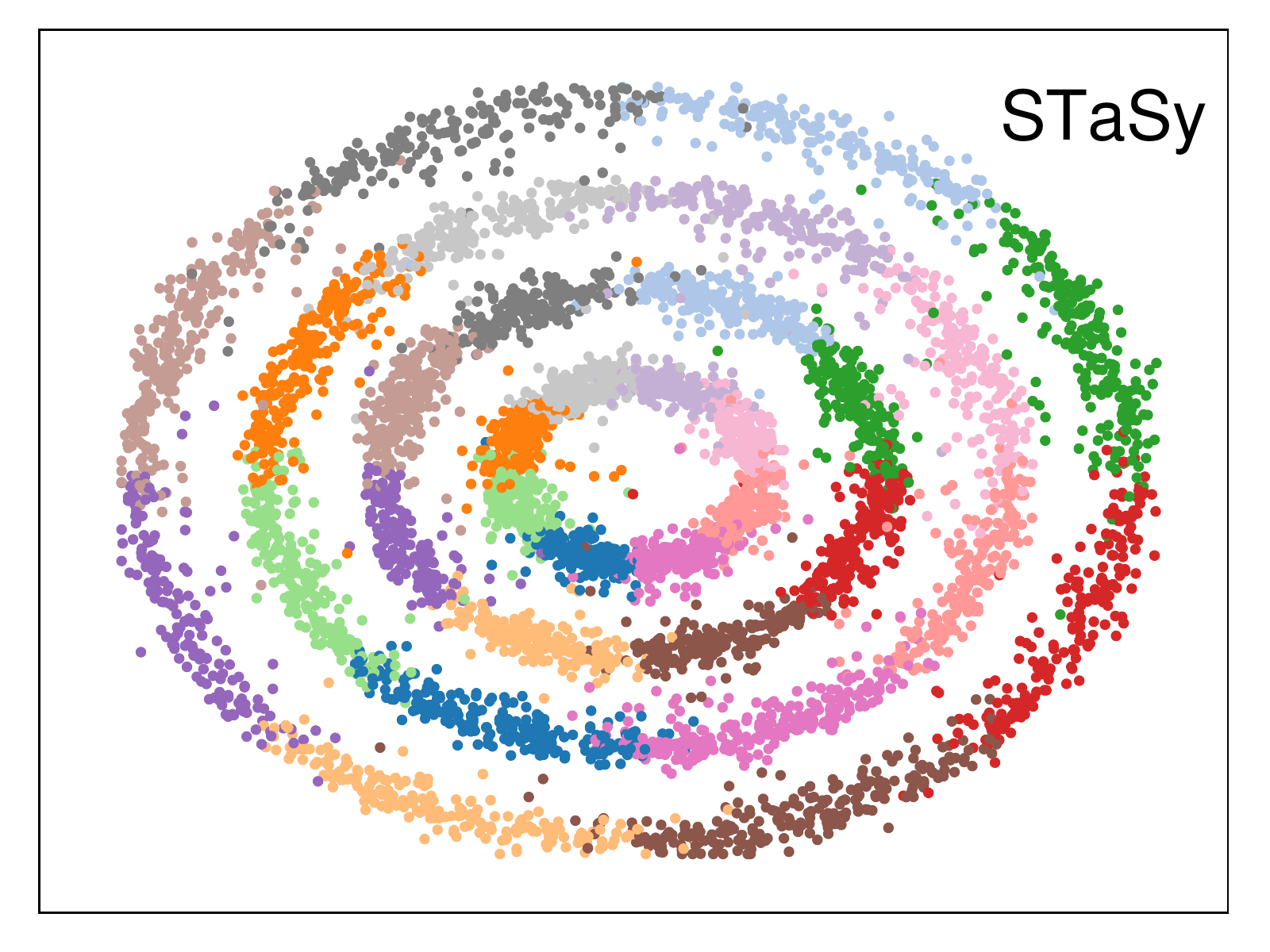}}
        \end{subfigure} 
       \begin{subfigure}{\includegraphics[width=0.25\columnwidth]{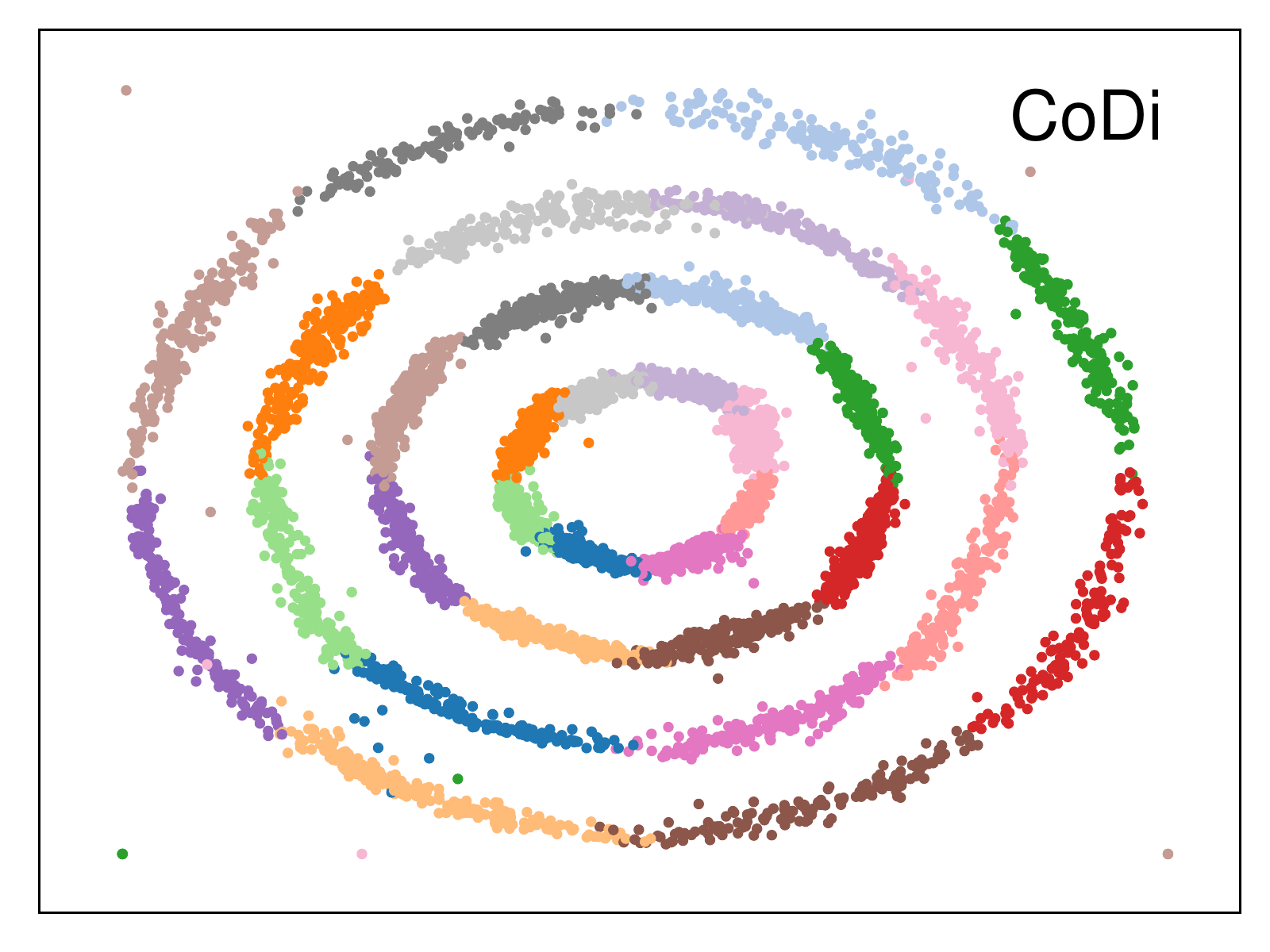}}
        \end{subfigure} 
        \caption{Preliminary experiment on a toy dataset}
        \label{fig:apd_toy}
\vspace{-1em}
\end{figure}
In Fig.~\ref{fig:apd_toy}, we provide detailed visualizations for the preliminary experiment on the toy dataset. The dataset contains 4 columns. The x-axis and y-axis are continuous variables, while the 16 colors and which circle the sample belongs to are discrete variables. As shown, fake data by \texttt{CoDi} is more similar to real data than fake data by \texttt{STaSy}. Specifically, fake data by \texttt{STaSy} fails to generate continuous and discrete variables precisely, resulting in noisy samples at the class boundaries.


\section{Network Architecture}\label{sec:net_archi}

\begin{figure}[h]
        \centering
        \includegraphics[trim={0cm 0cm 1.1cm 0cm},clip, width=1\columnwidth]{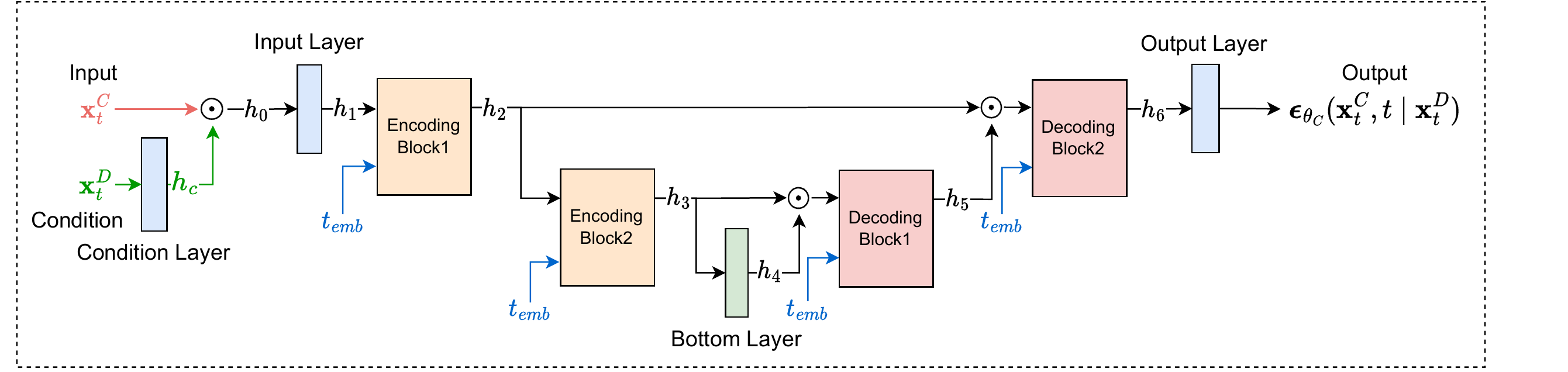}
        \caption{The continuous diffusion model network architecture diagram. The proposed continuous diffusion model network architecture is a modification of U-Net. We use the fully connected layers, and adopt the skip connection to pass the features from encoder to decoder. The continuous diffusion model is conditioned on discrete variables $\mathbf{x}_t^D$ and time $t$.}
        \label{fig:apd_archi}
        \vspace{-1em}
\end{figure}

Our proposed two diffusion model's network is U-Net-based architecture. We introduce an architecture only for the continuous diffusion model, but the same architecture is used for the discrete diffusion model. In Fig.~\ref{fig:apd_archi}, $\mathbf{x}_t^C$ is an input, $\mathbf{x}_t^D$ is a condition, $\odot$ is a concatenation operator, and $\boldsymbol{\epsilon}_{\theta_C}$ is an output of the continuous diffusion model. We use 4 layers; input, condition, bottom, and output layers, and 4 blocks; 2 encoding blocks and 2 decoding blocks linked together via a skip connection. The time $t$ is conditioned on the encoding and decoding blocks, and $t_{emb}$ is defined as follows:
\begin{equation}
    t_{emb} = \mathtt{FC}_{emb}^2(\mathtt{ReLU}(\mathtt{FC}_{emb}^1(\mathtt{Emb}(t)))),
\end{equation}
where $\mathtt{Emb}$ is a sinusoidal positional embedding~\cite{vaswani2017attention}, and $\mathtt{FC}$ is a fully connected layer. 
The condition layer inputs the condition $\mathbf{x}_t^D$ and outputs a hidden vector $h_c$, which has the dimension of half of the $\mathbf{x}_t^C$. 
Given the input $\mathbf{x}_t^C$ and the hidden vector $h_c$, we design the following continuous diffusion model $\boldsymbol{\epsilon}_{\theta_C}(\mathbf{x}_t^C, t\mid\mathbf{x}_t^D)$:
\begin{equation}
\begin{aligned}
h_{c} &= \mathtt{FC}_c(\mathbf{x}_t^D), \\
h_{0} &= \mathbf{x}_t^C\odot h_c ,\\
h_{1} &= \mathtt{FC}_1(h_{0}) , \\
h_{i} &= \mathtt{ReLU}(\mathtt{FC}_i^2(\mathtt{ReLU}(\mathtt{FC}_i^1(h_{i-1}))\oplus\mathtt{ReLU}(\mathtt{FC}_i^t(t_{emb})))), &\textrm{ if $i=2, 3,$} \\
h_{4} &= \mathtt{ReLU}(\mathtt{FC}_4(h_{3})), \\
h_{i} &= \mathtt{ReLU}(\mathtt{FC}_i^2(\mathtt{ReLU}(\mathtt{FC}_i^1(h_{i-1}\odot h_{8-i}))\oplus\mathtt{ReLU}(\mathtt{FC}_i^t(t_{emb})))), &\textrm{ if $i=5, 6,$} \\
\boldsymbol{\epsilon}_{\theta_C}(\mathbf{x}_t^C, t\mid\mathbf{x}_t^D) &= \mathtt{FC}_7(h_{6}) ,
\end{aligned}
\end{equation}
where  $\odot$ means the concatenation operator, $h_{i}$ is the $i^{th}$ hidden vector, and $\oplus$ means the element-wise addition.
\section{The Reverse Transition Probabilities for the Co-evolving Conditional Diffusion models}\label{sec:apd_proposition}

In Proposition~\ref{prop:cond}, we define the forward and reverse processes of the co-evolving conditional diffusion models.

The reverse transition probabilities of each reverse process are defined as follows:
\begin{equation}
p_{\theta_C}(\mathbf{x}_{t-1}^C|\mathbf{x}_{t}^C, \mathbf{x}_{t}^D) = \mathcal{N}(\mathbf{x}_{t-1}^C;\boldsymbol{\mu}_{\theta_C}(\mathbf{x}_t^C, t\mid\mathbf{x}_t^D), \sigma_t^2\mathbf{I}),
\end{equation}
\begin{equation}
p_{\theta_D}(\mathbf{x}_{t-1}^{D_i}|\mathbf{x}_{t}^{D_i}, \mathbf{x}_{t}^C)=\mathcal{C}(\mathbf{x}_{t-1}^{D_i};\mathbf{p}^i_{\theta_D}),
\end{equation}
where $1\leq i \leq N_D$, and $\boldsymbol{\mu}_{\theta_C}(\mathbf{x}_t^C, t\mid\mathbf{x}_t^D)$, $\sigma_t^2$, and $\mathbf{p}^i_{\theta_D}$ are defined as follows:
\begin{equation}
\boldsymbol{\mu}_{\theta_C}(\mathbf{x}_t^C, t\mid\mathbf{x}_t^D) = \frac{1}{\sqrt{\alpha_t}}(\mathbf{x}_t^C-\frac{\beta_t}{\sqrt{1-\bar{\alpha}_t}}\boldsymbol{\epsilon}_{\theta
_C}(\mathbf{x}_t^C, t\mid\mathbf{x}_t^D)),
\end{equation}
\begin{equation}
\sigma_t^2 = \frac{1-\bar{\alpha}_{t-1}}{1-\bar{\alpha}_t}\beta_t\mathbf{I},
\end{equation}
\begin{equation}
\mathbf{p}^i_{\theta_D}=\sum_{\hat{\mathbf{x}}_0^{D_i}=1}^{K_i}q(\mathbf{x}_{t-1}^{D_i}|\mathbf{x}_t^{D_i},\hat{\mathbf{x}}_0^{D_i})p_{\theta_D}(\hat{\mathbf{x}}_0^{D_i}|\mathbf{x}_t^{D_i}, \mathbf{x}_t^{C}).
\end{equation}
\section{Detailed Experimental Settings for Reproducibility}\label{sec:apd_env}

\subsection{Experimental Environments}
Our software and hardware environments are as follows: \textsc{Ubuntu} 18.04.6 LTS, \textsc{Python} 3.10.8, \textsc{Pytorch} 1.11.0, \textsc{CUDA} 11.7, and \textsc{NVIDIA} Driver 470.161.03, i9, CPU, and \textsc{NVIDIA RTX 3090}. 

\subsection{Hyperparameter Settings for \texttt{CoDi}}

Hyperparameter settings for the best models are summarized in Tables~\ref{tbl:hyperparameter1} and~\ref{tbl:hyperparameter2}. We use a learning rate in \{2e-03, 2e-05\}. We search for $\dim(\mathtt{Emb}(t))$, which is a time embedding dimension for every block consisting of network, in \{16, 32, 64, 128\} for each diffusion model. $\dim(h_1), \dim(h_2),$ and $\dim(h_3)$ decide the number of learnable parameters for networks, where $\dim(h_1)=\dim(h_6)$, $\dim(h_2)=\dim(h_5)$, and $\dim(h_3)=\dim(h_4)$, allowing the encoder and decoder part to be symmetry. We search for \{$\dim(h_1), \dim(h_2), \dim(h_3)$\} in \{\{16, 32, 64\}, \{32, 64, 128\}, \{64, 128, 256\}, \{128, 256, 512\}\}, considering the dataset size. The constrastive learning loss coefficient $\lambda_C$ and $\lambda_D$ are \{0.2, 0.3, \dots, 0.7, 0.8\}. We use a linear noise schedule for $\beta_t$, which is linearly increased from 1e-05 to 2e-02, and use the total diffusion timesteps as $T=50$.

\begin{table}[h]
\caption{The best hyperparameters used in Tables~\ref{table:samplingquality},~\ref{table:samplingdiversity}, and~\ref{table:samplingtime} for the continuous diffusion model}
\label{tbl:hyperparameter1}
\vskip 0.1in
\begin{center}
\begin{sc}
\begin{tabular}{lccccccccc}
\toprule
        
\multirow{2}{*}{Datasets} & \multicolumn{4}{c}{Continuous diffusion model} \\ 
\cmidrule{2-5} 
& Learning rate & $\dim(\mathtt{Emb}(t))$ & $ \dim(h_1), \dim(h_2), \dim(h_3) $ & $ \lambda_C $\\ 
\midrule

\texttt{Bank} & 2e-03 & 32 & 64,128,256 & 0.2  \\ 
\texttt{Heart} & 2e-03 & 16 & 64,128,256 & 0.2 \\ 
\texttt{Seismic} & 2e-03 & 64 & 64,128,256 & 0.4  \\
\texttt{Stroke} & 2e-05 & 16 & 64,128,256 & 0.2 \\
\texttt{CMC} & 2e-03 & 32 & 16,32,64 & 0.6 \\
\texttt{Customer} & 2e-03 & 16 & 32,64,128 & 0.2\\
\texttt{Faults} & 2e-03 & 32 & 64,128,256 & 0.2 \\
\texttt{Obesity} & 2e-03 & 64 & 64,128,256 & 0.5  \\
\texttt{Absent} & 2e-03 & 32 & 64,128,256 & 0.3  \\
\texttt{Drug} & 2e-03 & 32 & 64,128,256 & 0.2  \\
\texttt{Insurance} & 2e-03 & 32 & 64,128,256 & 0.2  \\

\bottomrule
\end{tabular}
\end{sc}
\end{center}
\vskip -0.1in
\end{table}

\begin{table}[h]
\caption{The best hyperparameters used in Tables~\ref{table:samplingquality},~\ref{table:samplingdiversity}, and~\ref{table:samplingtime} for the discrete diffusion model}
\label{tbl:hyperparameter2}
\vskip 0.1in
\begin{center}
\begin{sc}
\begin{tabular}{lccccccccc}
\toprule
        
\multirow{2}{*}{Datasets}  & \multicolumn{4}{c}{Discrete diffusion model} \\ 
\cmidrule{2-5} 
& Learning rate & $\dim(\mathtt{Emb}(t))$ & $ \dim(h_1), \dim(h_2), \dim(h_3) $ & $ \lambda_D $ \\ 
\midrule

\texttt{Bank} & 2e-03 & 64 & 64,128,256 & 0.3 \\ 
\texttt{Heart} & 2e-03 & 64 & 64,128,256 & 0.2 \\ 
\texttt{Seismic} & 2e-03 & 64 & 64,128,256 & 0.6 \\
\texttt{Stroke} & 2e-03 & 128 & 128,256,512 & 0.3 \\
\texttt{CMC}  & 2e-03 & 32 & 32,64,128 & 0.8 \\
\texttt{Customer}  & 2e-03 & 64 & 32,64,128 & 0.5 \\
\texttt{Faults}  & 2e-03 & 64 & 64,128,256 & 0.2 \\
\texttt{Obesity}  & 2e-03 & 64 & 64,128,256 & 0.3 \\
\texttt{Absent}  & 2e-05 & 32 & 64,128,256 & 0.2 \\
\texttt{Drug} & 2e-03 & 32 & 64,128,256 & 0.2 \\
\texttt{Insurance}  & 2e-03 & 32 & 128,256,512 & 0.2 \\

\bottomrule
\end{tabular}
\end{sc}
\end{center}
\vskip -0.1in
\end{table}

\subsection{Datasets}\label{sec:apd_dataset}
In this section, we provide the real-world datasets used for our experiments. The datasets are selected considering the number of continuous and discrete variables consisting of the tabular data. Table~\ref{tab:dataset} summarizes the statistical information of the datasets.

\begin{table}[h]
\caption{Dataset information used in our experiments. \#Train and \#Test are the numbers of training data and test data, respectively, and $N_C$ and $N_D$ are the numbers of continuous and discrete columns, respectively.}
\label{tab:dataset}
\vskip 0.1in
\begin{center}
\setlength\tabcolsep{4pt}
\begin{small}
\begin{sc}
\begin{tabular}{llcccc}
\toprule
Task & Datasets & \#Train & \#Test & $N_C$& $N_D$  \\
\midrule
\multirow{5}{*}{Binary} &\texttt{Bank} & 36169 & 9042 & 7 & 10 \\
&\texttt{Heart}  & 815 & 203 & 4 & 10 \\
&\texttt{Seismic} & 2068 & 516 & 11 & 5 \\
&\texttt{Stroke}   & 2740 & 685 & 2 & 8 \\
&\texttt{Phishing}   & 8844 & 2211 & 0 & 31\\
\midrule
\multirow{7}{*}{Multi-class}&\texttt{CMC}   & 1179 & 294 & 2 & 8 \\
&\texttt{Customer}  & 800 & 200 & 5 & 7 \\
&\texttt{Faults}   & 1553 &  388 & 24 & 4 \\
&\texttt{Obesity}   & 1689 &  422 & 8 & 9 \\
&\texttt{Car} & 1383 & 345 & 0& 7 \\
&\texttt{Clave}  & 8639 & 2159 & 0 & 17 \\
&\texttt{Nursery} & 10368 & 2592 & 0 & 9\\
\midrule
\multirow{3}{*}{Regression}&\texttt{Absent}   & 592 &  148 &  12 & 9 \\
&\texttt{Drug}   & 829 & 207 & 5 & 1 \\
&\texttt{Insurance}   & 789 & 197 &  3 & 8 \\
\bottomrule
\end{tabular}
\end{sc}
\end{small}
\end{center}
\vskip -0.1in
\end{table}
\begin{itemize}
    \item \texttt{Bank}~\cite{bank} is to predict whether a client subscribed a term deposit or not. The dataset contains personal financial situations, e.g., whether the person has housing loan or not.  (\url{https://archive.ics.uci.edu/ml/datasets/bank+marketing}) (CC BY 4.0)
    \item \texttt{Heart} is to predict the presence of heart disease in a patient, which consists of personal health condition. (\url{https://www.kaggle.com/datasets/johnsmith88/heart-disease-dataset}) (CC BY 4.0)
    \item \texttt{Seismic}~\cite{seismic} predicts the impact of an earthquake. The dataset contains seismic information such as the number of seismic bumps. (\url{https://archive.ics.uci.edu/ml/datasets/seismic-bumps}) (CC BY 4.0)
    \item \texttt{Stroke}~\cite{liu2019hybrid} is used to predict whether a patient is likely to get stroke based on the patient information. (CC0 1.0) (\url{https://www.kaggle.com/datasets/fedesoriano/stroke-prediction-dataset})
    \item \texttt{Phishing}~\cite{phishing} is to predict if a webpage is a phishing site. The dataset consists of important features for predicting the phishing sites, including information about webpage transactions. (\url{https://archive.ics.uci.edu/ml/datasets/phishing+websites}) (CC BY 4.0)
    \item \texttt{CMC}~\cite{cmc} is to predict the current contraceptive method chioce of a woman based on her demographic and socio-economic characteristics. (\url{https://archive.ics.uci.edu/ml/datasets/Contraceptive+Method+Choice}) (CC BY 4.0)
    \item \texttt{Customer} consists of information about the telecommunication company's customers and their groups. (\url{https://www.kaggle.com/prathamtripathi/customersegmentation}) (CC0 1.0)
    \item \texttt{Faults}~\cite{faults} is for fault detection in steel manufacturing process. (\url{https://archive.ics.uci.edu/ml/datasets/steel+plates+faults}) (CC BY 4.0)
    \item \texttt{Obesity}~\cite{obesity} is to estimate the obesity level based on eating habits and physical condition of individuals from Mexico, Peru, and Columbia. (\url{https://archive.ics.uci.edu/ml/datasets/Estimation+of+obesity+levels+based+on+eating+habits+and+physical+condition+}) (CC BY 4.0)
    \item \texttt{Car}~\cite{car} is to categorize a car condition, which contains car characteristics, such as the number of doors. (\url{https://archive.ics.uci.edu/ml/datasets/Car+Evaluation}) (CC BY 4.0)
    \item \texttt{Clave}~\cite{clave} is to predict the class to which input music belongs. The class is one of the \texttt{Neutral}, \texttt{Reverse Clave}, \texttt{Forward Clave}, and \texttt{Incoherent}. 
    (\url{https://archive.ics.uci.edu/ml/datasets/Firm-Teacher_Clave-Direction_Classification}) (CC BY 4.0)
    \item \texttt{Nursery}~\cite{nursery} is to rank applications for nursery schools. The dataset contains the family structures, parents' occupation, children's health condition, and so on. (\url{https://archive.ics.uci.edu/ml/datasets/nursery}) (CC BY 4.0)
    \item \texttt{Absent}~\cite{absent} predicts the absenteeism time in hours. The dataset consists demographic information. (\url{https://archive.ics.uci.edu/ml/datasets/Absenteeism+at+work}) (CC BY 4.0)
    \item \texttt{Drug}~\cite{drug} covers expenditure on prescription medicines and self-medication. (\url{https://data.oecd.org/healthres/pharmaceutical-spending.htm}) (CC BY-NC-SA 3.0 IGO)
    \item \texttt{Insurance} is for prediction on the yearly medical cover cost. The dataset contains a person's medical information. (\url{https://www.kaggle.com/datasets/tejashvi14/medical-insurance-premium-prediction}) (CC0 1.0)
    
\end{itemize}

 \subsection{Baselines}\label{sec:apd_baseline}
 We describe the baseline methods used in our experiment. The baseline methods are as follows, which consist of various types of generative models, from the GAN-based methods to the score-based generative model.
\begin{itemize}
    \item \texttt{MedGAN}~\cite{DBLP:journals/corr/ChoiBMDSS17} includes non-adversarial losses for discrete medical records. 
    \item \texttt{VEEGAN}~\cite{srivastava2017veegan} is a GAN equipped with a reconstructor network, which aims for diverse sampling. 
    \item \texttt{CTGAN} and \texttt{TVAE}~\cite{ctgan} handle challenges from the mixed type of variables. 
    \item \texttt{TableGAN}~\cite{tablegan} is a GAN for tabular data using convolutional neural networks. 
    \item \texttt{OCT-GAN}~\cite{octgan} contains neural networks based on neural ordinary differential equations. 
    \item \texttt{RNODE}~\cite{Finlay2020HowTT} is an advanced flow-based model for image data, and we customize the network to synthesize tabular data. 
    \item \texttt{STaSy}~\cite{stasy} is a currently proposed score-based generative model for tabular data synthesis.
\end{itemize}

\clearpage
\section{Additional Experimental Results}\label{sec:apd_fullresults}
\subsection{Sampling Quality}\label{sec:apd_quality}

We consider F1 and AUROC (resp. $R^2$ and RMSE) for the classification (resp. regression) tasks to evaluate the sampling quality. Full results for all datasets and baselines are in Tables~\ref{tab:apd_binary_quality} and~\ref{tab:apd_multi_quality} for classification, and in Table~\ref{tab:apd_reg_quality} for regression. For reliability, we report their mean and standard deviation of 5 evaluations on TSTR evaluation. \texttt{Identity} is a case of ``TRTR'', where we train classification/regression models with real training data and test them on real test data. As shown, \texttt{CoDi} outperforms the others in almost all cases. Especially in the regression task, \texttt{CoDi} is the only method that shows the positive $R^2$ in all datasets.

\begin{table}[h]
\vspace{-0.3em}
\caption{Binary classification results with real data. We report binary F1 and AUROC for binary classification. }
\label{tab:apd_binary_quality}
\vskip 0.1in
\begin{center}
\setlength\tabcolsep{2pt}
\begin{small}
\begin{sc}
\begin{tabular}{lccccccccccccc}
\toprule
\multirow{2}{*}{Methods}& \multicolumn{2}{c}{\texttt{Bank}} && \multicolumn{2}{c}{\texttt{Heart}} && \multicolumn{2}{c}{\texttt{Seismic}} && \multicolumn{2}{c}{\texttt{Stroke}}  \\ \cmidrule{2-3} \cmidrule{5-6} \cmidrule{8-9} \cmidrule{11-12}
& Binary F1 & AUROC && Binary F1 & AUROC && Binary F1 & AUROC && Binary F1 & AUROC \\ \midrule
        \texttt{Identity} & 0.520\scriptsize{±0.038} & 0.895\scriptsize{±0.080} && 0.972\scriptsize{±0.057} & 0.990\scriptsize{±0.025} && 0.102\scriptsize{±0.109} & 0.697\scriptsize{±0.055} && 0.068\scriptsize{±0.042} & 0.666\scriptsize{±0.085}  \\ \midrule 

        \texttt{MedGAN} & 0.000\scriptsize{±0.000} & 0.500\scriptsize{±0.000} && 0.594\scriptsize{±0.067} & 0.659\scriptsize{±0.061} && 0.000\scriptsize{±0.000} & 0.500\scriptsize{±0.000} && 0.015\scriptsize{±0.027} & 0.527\scriptsize{±0.022}  \\ 
        \texttt{VEEGAN} & 0.206\scriptsize{±0.015}  & 0.521\scriptsize{±0.061} && 0.525\scriptsize{±0.126} & 0.453\scriptsize{±0.114} && 0.161\scriptsize{±0.020}  & 0.590\scriptsize{±0.050}  && 0.144\scriptsize{±0.052} & \underline{0.644\scriptsize{±0.026}}  \\ 
        \texttt{CTGAN} & 0.515\scriptsize{±0.029} & 0.883\scriptsize{±0.022} && 0.611\scriptsize{±0.015} & 0.543\scriptsize{±0.040} && 0.088\scriptsize{±0.054} & 0.655\scriptsize{±0.061} && \underline{0.159\scriptsize{±0.022}} & 0.617\scriptsize{±0.028}  \\ 
        \texttt{TVAE} & 0.524\scriptsize{±0.019} & 0.870\scriptsize{±0.025} && 0.745\scriptsize{±0.056} & 0.843\scriptsize{±0.069} && 0.000\scriptsize{±0.000} & 0.500\scriptsize{±0.000} && 0.006\scriptsize{±0.015} & 0.534\scriptsize{±0.021}  \\ 
        \texttt{TableGAN} & 0.444\scriptsize{±0.060} & 0.812\scriptsize{±0.074} & &0.767\scriptsize{±0.023} & 0.868\scriptsize{±0.039} &&  0.219\scriptsize{±0.056} & 0.679\scriptsize{±0.043} && \cellcolor{cyan!15} \textbf{0.201\scriptsize{±0.015} }& 0.634\scriptsize{±0.019}  \\ 
        \texttt{OCT-GAN} & 0.539\scriptsize{±0.019} & 0.887\scriptsize{±0.020} && 0.649\scriptsize{±0.033} & 0.733\scriptsize{±0.056} && 0.209\scriptsize{±0.025} & 0.686\scriptsize{±0.077} && 0.128\scriptsize{±0.020} & 0.634\scriptsize{±0.041}  \\ 
        \texttt{RNODE} & 0.263\scriptsize{±0.043}  & 0.739\scriptsize{±0.083}  && 0.811\scriptsize{±0.037} & 0.895\scriptsize{±0.041} && 0.141\scriptsize{±0.044} & 0.515\scriptsize{±0.081} && 0.068\scriptsize{±0.011} & 0.511\scriptsize{±0.019}  \\ 
        \texttt{STaSy} & \underline{0.562\scriptsize{±0.022}} & \cellcolor{cyan!15} \textbf{0.905\scriptsize{±0.017}} &&\underline{0.835\scriptsize{±0.031}} & \underline{0.918\scriptsize{±0.025}} && \underline{0.295\scriptsize{±0.019}} & \underline{0.721\scriptsize{±0.032}} & &0.132\scriptsize{±0.018} & 0.641\scriptsize{±0.049}  \\ \midrule
        \texttt{CoDi} & \cellcolor{cyan!15} \textbf{0.566\scriptsize{±0.014}} & \underline{0.894\scriptsize{±0.023}} && \cellcolor{cyan!15} \textbf{0.872\scriptsize{±0.039} }& \cellcolor{cyan!15} \textbf{0.934\scriptsize{±0.038}} && \cellcolor{cyan!15} \textbf{0.305\scriptsize{±0.040}} & \cellcolor{cyan!15} \textbf{0.731\scriptsize{±0.036}} &&  0.147\scriptsize{±0.016} & \cellcolor{cyan!15} \textbf{0.684\scriptsize{±0.015}}  \\ 
\bottomrule
\end{tabular}
\end{sc}
\end{small}
\end{center}
\vskip -0.1in
\end{table}

\begin{table}[h]
\vspace{-0.3em}

\caption{Multi-class classification results with real data. We report macro F1 and AUROC for multi-class classification. }
\label{tab:apd_multi_quality}
\vskip 0.1in
\begin{center}
\setlength\tabcolsep{2pt}
\begin{small}
\begin{sc}
\begin{tabular}{lccccccccccccc}
\toprule
\multirow{2}{*}{Methods}& \multicolumn{2}{c}{\texttt{CMC}} && \multicolumn{2}{c}{\texttt{Customer}} && \multicolumn{2}{c}{\texttt{Faults}} && \multicolumn{2}{c}{\texttt{Obesity}}  \\ \cmidrule{2-3} \cmidrule{5-6} \cmidrule{8-9} \cmidrule{11-12}
& Macro F1 & AUROC && Macro F1 & AUROC && Macro F1 & AUROC && Macro F1 & AUROC \\ \midrule
        \texttt{Identity}&0.493\scriptsize{±0.009} & 0.709\scriptsize{±0.011} && 0.370\scriptsize{±0.022} & 0.669\scriptsize{±0.021} && 0.777\scriptsize{±0.052} & 0.923\scriptsize{±0.048} && 0.965\scriptsize{±0.014} & 0.992\scriptsize{±0.015}  \\ \midrule
        \texttt{MedGAN}&0.296\scriptsize{±0.015} & 0.513\scriptsize{±0.008} && 0.210\scriptsize{±0.043} & 0.493\scriptsize{±0.030} && 0.068\scriptsize{±0.000} & 0.500\scriptsize{±0.000} && 0.040\scriptsize{±0.000} & 0.500\scriptsize{±0.000}  \\ 
        \texttt{VEEGAN}&0.287\scriptsize{±0.036} & 0.533\scriptsize{±0.009} && 0.105\scriptsize{±0.000} & 0.500\scriptsize{±0.000} && 0.050\scriptsize{±0.000} & 0.500\scriptsize{±0.000} && 0.040\scriptsize{±0.000} & 0.500\scriptsize{±0.000}  \\ 
        \texttt{CTGAN}&0.335\scriptsize{±0.012} & 0.514\scriptsize{±0.017} && 0.247\scriptsize{±0.013} & 0.521\scriptsize{±0.011} && 0.221\scriptsize{±0.025} & 0.661\scriptsize{±0.073} && 0.139\scriptsize{±0.014} & 0.522\scriptsize{±0.024}  \\ 
        \texttt{TVAE}&0.297\scriptsize{±0.004} & 0.537\scriptsize{±0.026} && 0.140\scriptsize{±0.006} & 0.518\scriptsize{±0.005} && 0.068\scriptsize{±0.000} & 0.500\scriptsize{±0.000} && 0.439\scriptsize{±0.024} & 0.835\scriptsize{±0.016}  \\ 
        \texttt{TableGAN}&0.366\scriptsize{±0.016} & 0.582\scriptsize{±0.024} && 0.241\scriptsize{±0.007} & 0.512\scriptsize{±0.005} && 0.187\scriptsize{±0.017} & 0.593\scriptsize{±0.047} && 0.292\scriptsize{±0.008} & 0.741\scriptsize{±0.041}  \\ 
        \texttt{OCT-GAN}&0.345\scriptsize{±0.019} & 0.541\scriptsize{±0.034} && 0.227\scriptsize{±0.015} & 0.493\scriptsize{±0.003} && 0.357\scriptsize{±0.004} & 0.796\scriptsize{±0.049} && 0.396\scriptsize{±0.061} & 0.744\scriptsize{±0.085}  \\ 
        \texttt{RNODE}&0.394\scriptsize{±0.026} & 0.610\scriptsize{±0.038} && 0.315\scriptsize{±0.027} & 0.600\scriptsize{±0.036} && 0.298\scriptsize{±0.056} & 0.778\scriptsize{±0.085} && 0.469\scriptsize{±0.062} & 0.826\scriptsize{±0.068}  \\ 
        \texttt{STaSy}&\underline{0.492\scriptsize{±0.010}} & \underline{0.686\scriptsize{±0.027}} && \underline{0.337\scriptsize{±0.002}} & \cellcolor{cyan!15} \textbf{0.635\scriptsize{±0.044}} && \underline{0.668\scriptsize{±0.022}} & \underline{0.894\scriptsize{±0.039}} && \underline{0.900\scriptsize{±0.034}} & \underline{0.984\scriptsize{±0.019}}  \\ \midrule
        \texttt{CoDi} & \cellcolor{cyan!15} \textbf{0.503\scriptsize{±0.008}} & \cellcolor{cyan!15} \textbf{0.694\scriptsize{±0.013}} && \cellcolor{cyan!15} \textbf{0.352\scriptsize{±0.015}} & \underline{0.619\scriptsize{±0.036}} && \cellcolor{cyan!15} \textbf{0.715\scriptsize{±0.046}} & \cellcolor{cyan!15} \textbf{0.908\scriptsize{±0.040}} && \cellcolor{cyan!15} \textbf{0.919\scriptsize{±0.034}} & \cellcolor{cyan!15} \textbf{0.989\scriptsize{±0.012}}  \\ 
\bottomrule
\end{tabular}
\end{sc}
\end{small}
\end{center}
\vskip -0.1in
\end{table}

\begin{table}[!ht]
\vspace{-0.3em}
\caption{Regression results with real data. We report $R^2$ and RMSE for regression.}
\label{tab:apd_reg_quality}
\begin{center}
\setlength\tabcolsep{2pt}
\begin{small}
\begin{sc}
\begin{tabular}{lcccccccccc}
\toprule
\multirow{2}{*}{Methods}& \multicolumn{2}{c}{\texttt{Absent}} && \multicolumn{2}{c}{\texttt{Drug}} && \multicolumn{2}{c}{\texttt{Insurance}}  \\ \cmidrule{2-3} \cmidrule{5-6} \cmidrule{8-9} 
& $R^2$ & RMSE &&$R^2$ & RMSE && $R^2$ & RMSE  \\ \midrule
      \texttt{Identity}& 0.352\scriptsize{±0.072} & 0.795\scriptsize{±0.044} && 0.992\scriptsize{±0.009} & 0.128\scriptsize{±0.095} && 0.658\scriptsize{±0.317} & 0.155\scriptsize{±0.068}  \\ \midrule
      \texttt{MedGAN}& -inf & inf && -4.877\scriptsize{±1.800} & 4.170\scriptsize{±0.708} && -inf & inf  \\ 
      \texttt{VEEGAN}& -8.802\scriptsize{±5.721} & 2.912\scriptsize{±1.209} && -5.354\scriptsize{±3.583} & 4.195\scriptsize{±1.452} && -inf& inf  \\ 
      \texttt{CTGAN}& -inf & inf&& 0.034\scriptsize{±0.080} & 1.705\scriptsize{±0.072} && -0.196\scriptsize{±0.308} & 0.307\scriptsize{±0.038}  \\ 
      \texttt{TVAE}& -inf & inf && -inf & inf&& -inf& inf  \\ 
      \texttt{TableGAN}& -0.363\scriptsize{±0.116} & 1.152\scriptsize{±0.049} && 0.183\scriptsize{±0.141} & 1.567\scriptsize{±0.132} && -0.031\scriptsize{±0.290} & 0.285\scriptsize{±0.038}  \\ 
      \texttt{OCT-GAN}& \underline{-0.123\scriptsize{±0.064}} & \underline{1.046\scriptsize{±0.030}} && 0.019\scriptsize{±0.129} & 1.713\scriptsize{±0.115} && -0.156\scriptsize{±0.220} & 0.303\scriptsize{±0.027}  \\ 
      \texttt{RNODE}& -0.604\scriptsize{±0.285}  & 1.246\scriptsize{±0.108}  && -0.174\scriptsize{±0.712}  & 1.838\scriptsize{±0.127}  && -0.133\scriptsize{±0.160}  &  0.297\scriptsize{±0.084}  \\ 
      \texttt{STaSy}& -4.833\scriptsize{±2.768} & 2.303\scriptsize{±0.629} && \underline{0.568\scriptsize{±0.134}} & \underline{1.131\scriptsize{±0.181}} && \underline{0.306\scriptsize{±0.129}} & \underline{0.235\scriptsize{±0.022}}  \\
      \midrule
      \texttt{CoDi}& \cellcolor{cyan!15} \textbf{0.095\scriptsize{±0.022}} & \cellcolor{cyan!15} \textbf{0.940\scriptsize{±0.012}} && \cellcolor{cyan!15} \textbf{0.768\scriptsize{±0.049}} & \cellcolor{cyan!15} \textbf{0.832\scriptsize{±0.088}} && \cellcolor{cyan!15} \textbf{0.575\scriptsize{±0.398}} & \cellcolor{cyan!15} \textbf{0.171\scriptsize{±0.074}}  \\ 
\bottomrule
\end{tabular}
\end{sc}
\end{small}
\end{center}
\vspace{-5em}
\end{table}

\clearpage

\subsection{Sampling Diversity}\label{sec:apd_diversity}
To evaluate the sampling diversity of fake data, we consider coverage~\cite{coverage}. Full results are in Tables~\ref{tab:apd_diversity1},~\ref{tab:apd_diversity2} and~\ref{tab:apd_diversity3}. We measure the coverage 5 times with different fake data by each method and report their mean and standard deviation. \texttt{CoDi} shows comparable performance in terms of the sampling diversity to the original data in many cases.

\begin{table}[h]
\caption{Sampling diversity in terms of coverage on binary classification datasets}
\label{tab:apd_diversity1}
\vskip 0.1in
\begin{center}
\begin{small}
\begin{sc}
\begin{tabular}{lccccccc}
\toprule
Methods & \texttt{Bank} & \texttt{Heart} &\texttt{Seismic} & \texttt{Stroke}  \\ \midrule
        \texttt{MedGAN}&0.000\scriptsize{±0.000} & 0.090\scriptsize{±0.011} & 0.000\scriptsize{±0.000} & 0.010\scriptsize{±0.001}  \\ 
        \texttt{VEEGAN}&0.001\scriptsize{±0.000}  & 0.004\scriptsize{±0.000} & 	0.000\scriptsize{±0.000}  & 0.000\scriptsize{±0.000} \\ 
        \texttt{CTGAN}&0.774\scriptsize{±0.001} & 0.098\scriptsize{±0.010} & 0.341\scriptsize{±0.013} & 0.664\scriptsize{±0.015}  \\ 
        \texttt{TVAE}&\underline{0.783\scriptsize{±0.003}} & 0.287\scriptsize{±0.010} & 0.401\scriptsize{±0.009} & 0.610\scriptsize{±0.016}  \\ 
        \texttt{TableGAN}&0.623\scriptsize{±0.004} & 0.816\scriptsize{±0.031} & \cellcolor{cyan!15} \textbf{0.553\scriptsize{±0.011}} & \underline{0.896\scriptsize{±0.008}}  \\ 
        \texttt{OCT-GAN}&0.623\scriptsize{±0.004} & 0.004\scriptsize{±0.000} & 0.326\scriptsize{±0.015} & 0.594\scriptsize{±0.014}  \\ 
        \texttt{RNODE}& 0.403\scriptsize{±0.003}  & 0.679\scriptsize{±0.034} & 0.198\scriptsize{±0.010} & 0.251\scriptsize{±0.010}  \\ 
        \texttt{STaSy}&\cellcolor{cyan!15} \textbf{0.854\scriptsize{±0.012}} & \underline{0.839\scriptsize{±0.009}} & \underline{0.505\scriptsize{±0.017}} & 0.789\scriptsize{±0.021}  \\ \midrule
        \texttt{CoDi}&0.687\scriptsize{±0.002} & \cellcolor{cyan!15} \textbf{0.949\scriptsize{±0.012} }& 0.359\scriptsize{±0.005} & \cellcolor{cyan!15} \textbf{0.919\scriptsize{±0.008}} \\ 
\bottomrule
\end{tabular}
\end{sc}
\end{small}
\end{center}
\vskip -0.1in
\end{table}

\begin{table}[h]
\caption{Sampling diversity in terms of coverage on multi-class classification datasets }
\label{tab:apd_diversity2}
\vskip 0.1in
\begin{center}
\begin{small}
\begin{sc}
\begin{tabular}{lccccccc}
\toprule
Methods & \texttt{CMC} & \texttt{Customer} & \texttt{Faults} & \texttt{Obesity} \\ \midrule
        \texttt{MedGAN}& 0.012\scriptsize{±0.004} & 0.015\scriptsize{±0.001} &0.002\scriptsize{±0.000} & 0.000\scriptsize{±0.000}  \\ 
        \texttt{VEEGAN} & 0.000\scriptsize{±0.000} & 0.003\scriptsize{±0.000} &0.002\scriptsize{±0.000} & 0.000\scriptsize{±0.000} \\
        \texttt{CTGAN} & 0.715\scriptsize{±0.008} & 0.428\scriptsize{±0.013}&0.031\scriptsize{±0.005} & 0.275\scriptsize{±0.008} \\
        \texttt{TVAE}& 0.603\scriptsize{±0.014} & 0.398\scriptsize{±0.010} &0.149\scriptsize{±0.007} & 0.359\scriptsize{±0.010} \\
        \texttt{TableGAN}& 0.766\scriptsize{±0.018} & \cellcolor{cyan!15}\textbf{ 0.896\scriptsize{±0.013}} &\underline{0.205\scriptsize{±0.020}} & 0.348\scriptsize{±0.011} \\ 
        \texttt{OCT-GAN} & 0.303\scriptsize{±0.011} & 0.073\scriptsize{±0.009}&0.054\scriptsize{±0.007} & 0.299\scriptsize{±0.006} \\
        \texttt{RNODE}& 0.869\scriptsize{±0.015} & 0.723\scriptsize{±0.024} &0.138\scriptsize{±0.007} & 0.313\scriptsize{±0.008} \\
        \texttt{STaSy}& \cellcolor{cyan!15} \textbf{0.943\scriptsize{±0.007}} & 0.713\scriptsize{±0.025} & {0.202\scriptsize{±0.014}} & \underline{0.633\scriptsize{±0.007}} \\ \midrule
        \texttt{CoDi} & \underline{0.934\scriptsize{±0.015}} & \underline{0.833\scriptsize{±0.021}} &\cellcolor{cyan!15} \textbf{0.270\scriptsize{±0.017}} & \cellcolor{cyan!15} \textbf{0.742\scriptsize{±0.015}} \\
\bottomrule
\end{tabular}
\end{sc}
\end{small}
\end{center}
\vskip -0.1in
\end{table}

\begin{table}[h]
\caption{Sampling diversity in terms of coverage on regression datasets}
\label{tab:apd_diversity3}
\vskip 0.1in
\begin{center}
\begin{small}
\begin{sc}
\begin{tabular}{lccccc}
\toprule
Methods &\texttt{Absent} & \texttt{Drug} & \texttt{Insurance} \\ \midrule
        \texttt{MedGAN} & 0.015\scriptsize{±0.001} & 0.009\scriptsize{±0.002}& 0.018\scriptsize{±0.005}  \\ 
        \texttt{VEEGAN}& 0.002\scriptsize{±0.000} & 0.002\scriptsize{±0.000} & 0.008\scriptsize{±0.000}  \\ 
        \texttt{CTGAN}& 0.287\scriptsize{±0.012}  & 0.557\scriptsize{±0.018} & 0.048\scriptsize{±0.005} \\ 
        \texttt{TVAE}& 0.082\scriptsize{±0.009}  & 0.588\scriptsize{±0.018} & 0.034\scriptsize{±0.011} \\ 
        \texttt{TableGAN} & \underline{0.294\scriptsize{±0.017}}  & \cellcolor{cyan!15} \textbf{0.880\scriptsize{±0.019}} & 0.059\scriptsize{±0.018} \\ 
        \texttt{OCT-GAN} & 0.044\scriptsize{±0.007} & 0.460\scriptsize{±0.021}  & 0.023\scriptsize{±0.004} \\ 
        \texttt{RNODE} & 0.085\scriptsize{±0.007}  & 0.395\scriptsize{±0.018}  & 0.172\scriptsize{±0.014}  \\ 
        \texttt{STaSy}& 0.004\scriptsize{±0.003} & 0.662\scriptsize{±0.016} & \underline{0.206\scriptsize{±0.016}}  \\ \midrule
        \texttt{CoDi} & \cellcolor{cyan!15} \textbf{0.843\scriptsize{±0.023}}  & \underline{0.827\scriptsize{±0.046}} & \cellcolor{cyan!15} \textbf{0.262\scriptsize{±0.020}}  \\ 
\bottomrule
\end{tabular}
\end{sc}
\end{small}
\end{center}
\vskip -0.1in
\end{table}

\clearpage

\subsection{Sampling Time}\label{sec:apd_time}
Tables~\ref{tab:apd_time1},~\ref{tab:apd_time2} and~\ref{tab:apd_time3} summarize the sampling time evaluation results. We measure the wall-clock time taken to sample 10K fake records 5 times and report their mean and standard deviation. \texttt{TableGAN} and \texttt{TVAE} show fast sampling time in almost all cases, while \texttt{RNODE} takes the longest time. \texttt{CoDi} shows much faster sampling time compared to \texttt{STaSy}, which is state-of-the-art score-based generative model. 

\begin{table}[h]
\caption{We summarize the sampling time for each binary classification dataset and tabular data synthesis method.}
\label{tab:apd_time1}
\vskip 0.1in
\begin{center}
\begin{small}
\begin{sc}
\begin{tabular}{lccccccc}
\toprule
Methods & \texttt{Bank} & \texttt{Heart} &\texttt{Seismic} & \texttt{Stroke}   \\ \midrule
        \texttt{MedGAN}&0.019\scriptsize{±0.000} & 0.019\scriptsize{±0.000} & 0.019\scriptsize{±0.000} & 0.019\scriptsize{±0.000}   \\ 
        \texttt{VEEGAN}&\underline{0.016\scriptsize{±0.000}} & 0.018\scriptsize{±0.001} & 0.017\scriptsize{±0.001} & 0.017\scriptsize{±0.002}  \\ 
        \texttt{CTGAN}&0.138\scriptsize{±0.000} & 0.110\scriptsize{±0.001} & 0.138\scriptsize{±0.000} & 0.101\scriptsize{±0.001}  \\ 
        \texttt{TVAE}&\cellcolor{cyan!15}\textbf{0.014\scriptsize{±0.001}} & \underline{0.012\scriptsize{±0.000}} & \underline{0.011\scriptsize{±0.000}} & \underline{0.012\scriptsize{±0.000}}   \\ 
        \texttt{TableGAN}&0.058\scriptsize{±0.002} & \cellcolor{cyan!15}\textbf{0.005\scriptsize{±0.000} }& \cellcolor{cyan!15}\textbf{0.005\scriptsize{±0.000}} & \cellcolor{cyan!15}\textbf{0.007\scriptsize{±0.000}}  \\ 
        \texttt{OCT-GAN}&0.477\scriptsize{±0.001} & 0.467\scriptsize{±0.001} & 0.576\scriptsize{±0.000} & 0.829\scriptsize{±0.000}   \\ 
        \texttt{RNODE}&229.254\scriptsize{±3.280} & 95.677\scriptsize{±1.838} & 35.550\scriptsize{±0.839} & 230.209\scriptsize{±0.771}   \\
        \texttt{STaSy}&2.069\scriptsize{±0.004} & 7.501\scriptsize{±1.457} & 3.230\scriptsize{±0.150} & 3.357\scriptsize{±0.108} \\ \midrule
        \texttt{CoDi}&1.321\scriptsize{±0.022} & 0.600\scriptsize{±0.240} & 0.259\scriptsize{±0.014} & 0.793\scriptsize{±0.417}  \\ 
\bottomrule
\end{tabular}
\end{sc}
\end{small}
\end{center}
\vskip -0.1in
\end{table}

\begin{table}[h]
\caption{We summarize the sampling time for each multi-class classification dataset and tabular data synthesis method.}
\label{tab:apd_time2}
\vskip 0.1in
\begin{center}
\begin{small}
\begin{sc}
\begin{tabular}{lccccccc}
\toprule
Methods &  \texttt{CMC} & \texttt{Customer} & \texttt{Faults} & \texttt{Obesity}  \\ \midrule
        \texttt{MedGAN} & 0.019\scriptsize{±0.000} & 0.030\scriptsize{±0.000}&\underline{0.019\scriptsize{±0.000}} & 0.019\scriptsize{±0.001}  \\ 
        \texttt{VEEGAN}& 0.016\scriptsize{±0.001} & 0.015\scriptsize{±0.000}&{0.022\scriptsize{±0.001}} & \underline{0.018\scriptsize{±0.001}}   \\ 
        \texttt{CTGAN}& 0.125\scriptsize{±0.000} & 0.134\scriptsize{±0.000}  &0.158\scriptsize{±0.000} & 0.126\scriptsize{±0.006}  \\ 
        \texttt{TVAE}& \underline{0.011\scriptsize{±0.000}} & \underline{0.011\scriptsize{±0.000}} &\cellcolor{cyan!15}\textbf{0.013\scriptsize{±0.000}} & \cellcolor{cyan!15}\textbf{0.014\scriptsize{±0.001} } \\ 
        \texttt{TableGAN}& \cellcolor{cyan!15}\textbf{0.007\scriptsize{±0.000}} & \cellcolor{cyan!15}\textbf{0.006\scriptsize{±0.000}} &0.029\scriptsize{±0.000} & 0.059\scriptsize{±0.002} \\ 
        \texttt{OCT-GAN} & 0.744\scriptsize{±0.329} & 0.606\scriptsize{±0.070}&0.462\scriptsize{±0.008} & 0.457\scriptsize{±0.002}  \\ 
        \texttt{RNODE} & 32.930\scriptsize{±0.279} & 36.931\scriptsize{±0.394} &52.887\scriptsize{±1.810} & 85.399\scriptsize{±0.542} \\
        \texttt{STaSy}&  1.952\scriptsize{±0.007} & 5.111\scriptsize{±0.224} &2.063\scriptsize{±0.024} & 2.051\scriptsize{±0.027}  \\ \midrule
        \texttt{CoDi}&  0.306\scriptsize{±0.019} & 0.290\scriptsize{±0.019} &0.241\scriptsize{±0.012} & 0.403\scriptsize{±0.044} \\ 
\bottomrule
\end{tabular}
\end{sc}
\end{small}
\end{center}
\vskip -0.1in
\end{table}
\begin{table}[!ht]
\caption{We summarize the sampling time for each regression dataset and tabular data synthesis method.}
\label{tab:apd_time3}
\vskip 0.1in
\begin{center}
\begin{small}
\begin{sc}
\begin{tabular}{lccccccc}
\toprule
Methods &\texttt{Absent} & \texttt{Drug} & \texttt{Insurance} \\ \midrule
        \texttt{MedGAN} & \cellcolor{cyan!15}\textbf{0.020\scriptsize{±0.001}} & 0.019\scriptsize{±0.000} & 0.019\scriptsize{±0.000}  \\ 
        \texttt{VEEGAN}& \underline{0.021\scriptsize{±0.001}} & 0.013\scriptsize{±0.001} & \underline{0.014\scriptsize{±0.000}}  \\ 
        \texttt{CTGAN}& 0.149\scriptsize{±0.001} & 0.103\scriptsize{±0.000} & 0.104\scriptsize{±0.000}  \\
        \texttt{TVAE}& 0.023\scriptsize{±0.000} & \underline{0.012\scriptsize{±0.000}} & 0.021\scriptsize{±0.000}  \\ 
        \texttt{TableGAN} & 0.061\scriptsize{±0.001} & \cellcolor{cyan!15}\textbf{0.005\scriptsize{±0.000}} & \cellcolor{cyan!15}\textbf{0.005\scriptsize{±0.000}}  \\ 
        \texttt{OCT-GAN} & 0.583\scriptsize{±0.005} & 0.703\scriptsize{±0.000} & 0.704\scriptsize{±0.000}  \\ 
        \texttt{RNODE} & 220.460\scriptsize{±1.814} & 43.188\scriptsize{±0.480} & 72.110\scriptsize{±0.331}  \\ 
        \texttt{STaSy}& 16.908\scriptsize{±0.062} & 2.965\scriptsize{±0.225} & 3.849\scriptsize{±3.015}  \\ \midrule
        \texttt{CoDi} & 0.764\scriptsize{±0.383} & 0.269\scriptsize{±0.016} & 0.460\scriptsize{±0.001}  \\ 
\bottomrule
\end{tabular}
\end{sc}
\end{small}
\end{center}
\vskip -0.1in
\end{table}

\end{document}